\documentclass[runningheads]{llncs}
\usepackage{graphicx}
\usepackage{amsmath,amssymb} 
\usepackage{color}

\usepackage{algorithm}
\usepackage{algpseudocode}
\usepackage{cite}

\usepackage{array}
\newcolumntype{L}[1]{>{\raggedright\let\newline\\\arraybackslash\hspace{0pt}}m{#1}}
\newcolumntype{C}[1]{>{\centering\let\newline\\\arraybackslash\hspace{0pt}}m{#1}}
\newcolumntype{R}[1]{>{\raggedleft\let\newline\\\arraybackslash\hspace{0pt}}m{#1}}

\usepackage{booktabs}
\usepackage[caption=false]{subfig}

\begin{document}
\title{Fast Semantic Segmentation on Video Using Block Motion-Based Feature Interpolation}

\titlerunning{Fast Semantic Segmentation on Video Using Block Motion}
%
\author{Samvit Jain \and
Joseph E. Gonzalez}
%
\authorrunning{S. Jain and J. E. Gonzalez}
%

\institute{University of California, Berkeley \\
\email{\{samvit,jegonzal\}@eecs.berkeley.edu}}
\maketitle              
\begin{abstract}
Convolutional networks optimized for accuracy on challenging, dense prediction tasks are prohibitively slow to run on each frame in a video. The spatial similarity of nearby video frames, however, suggests opportunity to reuse computation. Prior work has explored basic feature reuse and feature warping based on optical flow, but has encountered limits to the speedup attainable with these techniques. In this paper, we present a new, two part approach to accelerating inference on video. First, we propose a fast feature propagation technique that utilizes the block motion vectors present in compressed video (e.g. H.264 codecs) to cheaply propagate features from frame to frame. Second, we develop a novel feature estimation scheme, termed feature interpolation, that fuses features propagated from enclosing keyframes to render accurate feature estimates, even at sparse keyframe frequencies. We evaluate our system on the Cityscapes and CamVid datasets, comparing to both a frame-by-frame baseline and related work. We find that we are able to substantially accelerate segmentation on video, achieving \textit{near real-time frame rates} (20.1 frames per second) on large images ($960 \times 720$ pixels), while maintaining competitive accuracy. This represents an improvement of almost $6\times$ over the single-frame baseline and $2.5\times$ over the fastest prior work.

\end{abstract}
\section{Introduction}

\begin{figure}[t]
	\centering
	\includegraphics[width=9.5cm]{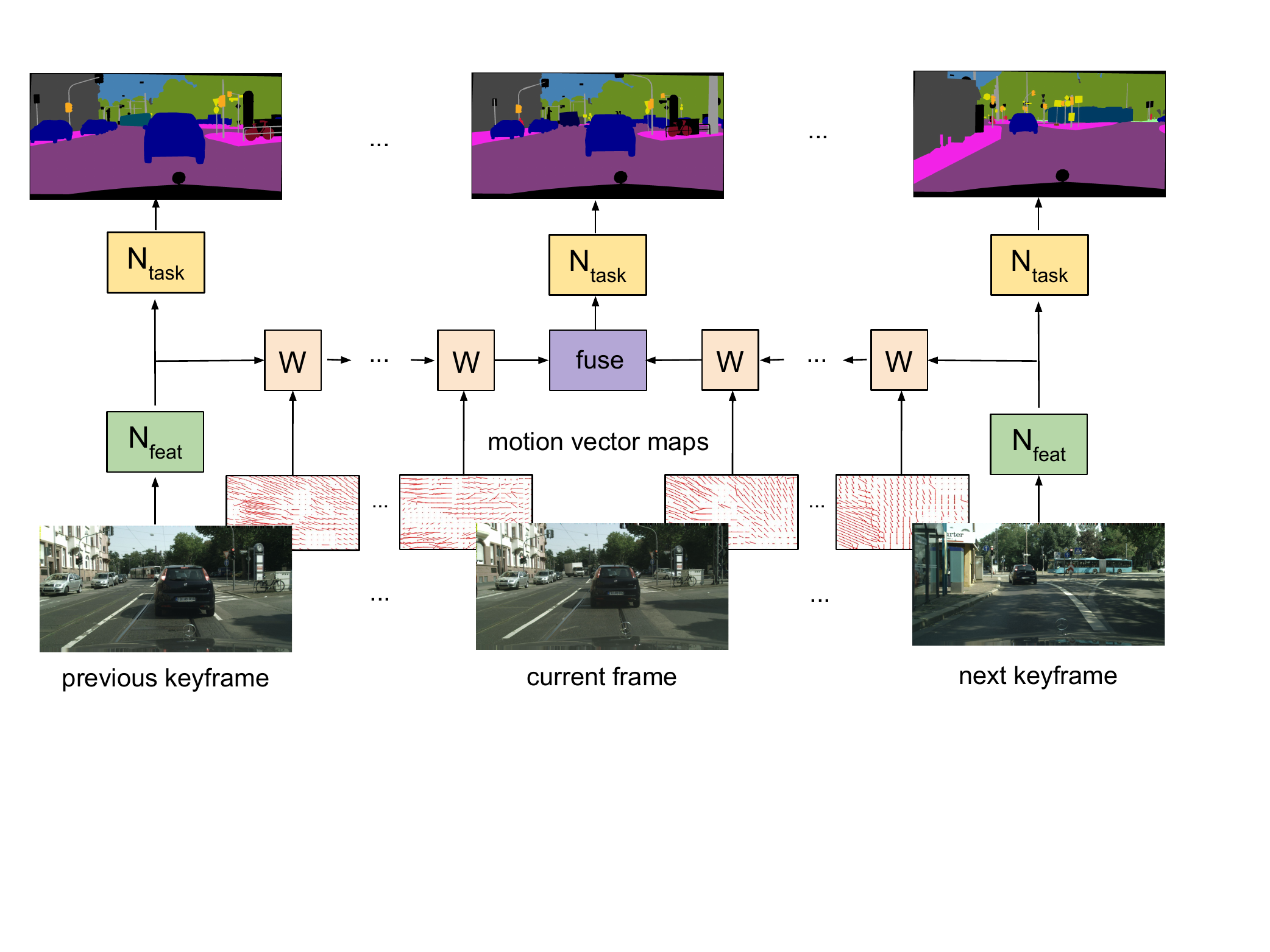}
	\vspace{-21mm}
	\caption{Feature interpolation warps ($W$) and fuses the features of enclosing keyframes to generate accurate feature estimates for intermediate frames.}
	\label{fig:abstract}
	\vspace{-4mm}
\end{figure}

Semantic segmentation, the task of assigning each pixel in an image to a semantic object class, is a problem of long-standing interest in computer vision. Since the first paper to suggest the use of fully convolutional networks to segment images \cite{FCN}, increasingly sophisticated architectures have been proposed, with the goal of segmenting more complex images, from larger, more realistic datasets, at higher accuracy \cite{SegNet, Multiscale, RefineNet, DeepLabV1, PSPNet, DeepLab}. The result has been a ballooning in both model size and inference times, as the core feature networks, borrowed from image classification models, have grown in layer depth and parameter count, and as the cost of a forward pass through the widest convolutional layers, a function of the size and detail of the input images, has risen in step. As a result, state-of-the-art networks today require between 0.5 to 3.0 seconds to segment a \textit{single}, high-resolution image (e.g. $2048 \times 1024$ pixels) at competitive accuracy \cite{DFF, NetWarp}.

At the same time, a new target data format for semantic segmentation has emerged: video. The motivating use cases include both batch settings, where video is segmented in bulk to generate training data for other models (e.g. autonomous control systems), and streaming settings, where high-throughput video segmentation enables interactive analysis of live footage (e.g. at surveillance sites). Video here consists of long sequences of images, shot at high frame rates (e.g. 30 frames per second) in complex environments (e.g. urban cityscapes) on modern, high-definition cameras (i.e. multi-megapixel). Segmenting individual frames at high accuracy still calls for the use of competitive image segmentation models, but the inference cost of these networks precludes their naive deployment on every frame in a multi-hour raw video stream.

A defining characteristic of realistic video is its high level of temporal continuity. Consecutive frames demonstrate significant spatial similarity, which suggests the potential to reuse computation across frames. Building on prior work, we exploit two observations: 1) higher-level features evolve more slowly than raw pixel content in video, and 2) feature computation tends to be much more expensive than task computation across a range of vision tasks (e.g. object detection, semantic segmentation) \cite{CC, DFF}. Accordingly, we divide our semantic segmentation model into a deep feature network and a cheap, shallow task network \cite{DFF}. We compute features only on designated keyframes, and propagate them to intermediate frames, by warping the feature maps with a frame-to-frame motion estimate. The task network is executed on all frames. Given that feature warping and task computation is much cheaper than feature extraction, a key parameter we aim to optimize is the interval between designated keyframes.

Here we make two key contributions to the effort to accelerate semantic segmentation on video. Firstly, noting the high level of data redundancy in video, we successfully utilize an artifact of compressed video, block motion vectors, to cheaply propagate features from frame to frame. Unlike other motion estimation techniques, which introduce extra computation on intermediate frames, block motion vectors are freely available in modern video formats, making for a simple, fast design. Secondly, we propose a novel feature estimation scheme that enables the features for a large fraction of the frames in a video to be inferred accurately and efficiently (see Fig. \ref{fig:abstract}). The approach works as follows: when computing the segmentation for a keyframe, we also precompute the features for the \textit{next} designated keyframe. Features for all subsequent intermediate frames are then computed as a \textit{fusion} of features warped forward from the last visited keyframe, and features warped backward from the incoming keyframe. This procedure thus implements an \textit{interpolation} of the features of the two closest keyframes.

We evaluate our framework on the Cityscapes and CamVid datasets. Our baseline consists of running a state-of-the-art segmentation network, DeepLab \cite{DeepLab}, on every frame, a setup that achieves published accuracy \cite{DCN}, and a throughput of 1.3 frames per second (fps) on Cityscapes and 3.6 fps on CamVid. Our improvements come in two phases. Firstly, our use of block motion vectors for feature propagation allow us to cut inference time on intermediate frames by 53\%, compared to approaches based on optical-flow, such as \cite{DFF}. Second, our bi-directional feature warping and fusion scheme enables substantial accuracy improvements, especially at high keyframe intervals. Together, the two techniques allow us to operate at \textit{over twice the average inference speed} as the fastest prior work, at any target level of accuracy. For example, if we are willing to tolerate no worse than 65 mIoU on our CamVid video stream, we are able to operate at a throughput of 20.1 fps, compared to the 8.0 fps achieved by the forward flow-based propagation from \cite{DFF}, a speedup of $2.5\times$. Overall, even when operating in high accuracy regimes (e.g. within 3\% mIoU of the baseline), we are able to accelerate segmentation on video by a factor of $2$-$6\times$.

\section{Related Work}

\subsubsection{Fast Video Inference.} Prior work on accelerating video segmentation focuses on reducing the frequency of feature computations across frames. Schemes have been proposed to reuse cached feature computations from previous frames as is \cite{CC}, and to propagate features forward from designated keyframes via warping with optical flow estimates (Zhu, et al., \cite{DFF}). While the latter approach arguably supersedes the former by allowing feature maps to evolve across frames, it suffers from one key shortcoming: features can be warped forward only as long as frame-to-frame changes consist primarily of internal displacements. Other forms of temporal evolution, such as the appearance of new objects, and simple perspective changes, such as camera pans, render past feature maps obsolete. In settings with complex dynamics, such as the urban environments captured in the Cityscapes and CamVid datasets \cite{Cityscapes, CamVid}, or in footage with fast ego motion, features must be recomputed frequently, limiting the attainable speedup.

Recent work also explores alternatives to feature reuse. Mahasseni et al. use an LSTM network to select designated frames for full segmentation \cite{BudgetAware}, and propagate final labels to other frames \cite{ActiveFrame}. Zhu et al. augment their earlier work in \cite{DFF2}, and look at feature estimate repair and adaptive keyframe scheduling. Using temporal information in video object detection and segmentation is also well-studied \cite{TCNN, FSO}, but not in the context of reducing inference cost.

\subsubsection{Feature Fusion.} Karpathy, et al. discuss various fusion schemes in the context of video classification \cite{Karpathy}, delineating early, late, and slow fusion approaches based on the stage at which information is merged. Fusion is implemented by stacking input frames, and adding a fourth, temporal dimension to convolutional filters. Another body of work, starting with \cite{TwoStream14}, studies spatial and temporal two-stream fusion for video action recognition \cite{TwoStream16, STResNets, DeepReps}. Feichtenhofer, et al. look specifically at fusing feature maps in \cite{TwoStream16}, and compare various spatial fusion techniques: sum fusion, max fusion, concatenation fusion, bilinear fusion, and convolutional fusion. They note both accuracy and parameter count-related tradeoffs, observations that broadly inform the choice of fusion strategies we consider. Recently, Jain et al. apply the two-stream model to video object segmentation, fusing appearance (i.e. frame-level) and motion (i.e. optical flow) streams to segment objects in weakly annotated videos \cite{FusionSeg}. In this paper, we diverge from the two-stream model, and instead propose the application of basic feature fusion schemes, as in \cite{TwoStream16}, to the problem of fast feature estimation.

\subsubsection{Motion in Video.} 

Fischer, et al. designed convolutional networks (FlowNet) to estimate optical flow \cite{FlowNet, FlowNet2}. Their models are used to generate optical flow maps for feature warping in \cite{DFF}, and can be jointly trained with other components of the video segmentation network. Other commonly studied motion estimation schemes include point tracking \cite{PointTrack} and scene flow estimation \cite{SceneFlow}, but these demonstrate less relevance to the problem of warping 2D feature maps for semantic segmentation. In general, motion information has been used extensively to track and segment objects in videos \cite{FastObjSeg, MovingObjects, FewStrokes, ObjectFlow}.

Recent work by Wu, et al. proposes the idea of training directly on compressed video to improve both accuracy and performance on video action recognition \cite{CoViAR}. Unlike \cite{CoViAR}, we use compressed video artifacts to infer motion for feature warping, not to reduce the size of input data. Moreover, our focus on pixel-level, dense prediction tasks, as opposed to video-level tasks \cite{CoViAR, MVecCNNs}, places our work in a different space, one which requires rendering predictions for each uncompressed video frame and which calls in particular for inference speedups.

\section{System Overview}

\subsection{Network architecture}

For our semantic segmentation network, we adhere to the common practice of adapting a competitive image classification architecture (e.g. ResNet-101) into a fully convolutional network capable of outputting class predictions for each pixel in the input image (e.g. DeepLab) \cite{FCN, DRN, DeepLab}. We identify two logical components in our final network: a \textit{feature network}, which takes as input an image $i \in R^{1 \times 3 \times h \times w}$ and outputs a representation $f_i \in R^{1 \times A \times \frac{h}{16} \times \frac{w}{16}}$, and a \textit{task network}, which given the representation, computes class predictions for each pixel in the image, $p_i \in R^{1 \times C \times h \times w}$. 

The feature network is obtained by eliminating the final, $k$-way classification layer in the chosen image classification architecture. The task network is built by concatenating: 1) a $1 \times 1$ convolutional feature projection layer, which reduces the feature channel dimensionality from $A$ to $\frac{A}{2}$, 2) a non-linear activation (e.g. a ReLU), 3) a $1 \times 1$ convolutional scoring layer, with channel dimension $C$, that outputs scores for each of $C$ classes in the target dataset, 4) a deconvolutional layer, which bilinearly upsamples the score maps to the resolution of the original image, and 5) a softmax layer, which converts scores to normalized probabilities for each pixel and object class.

We divide our network into these two functional components for two reasons. Firstly, feature maps are transferable. As  \cite{CC} observes, high-level feature maps evolve more slowly than raw pixels in a video. As a result, propagated features can serve as a good estimate for the features of proximate frames. Moreover, the fact that our feature outputs retain significant spatial structure means that we can often do better than mere copying, using estimates of frame-to-frame motion to instead \textit{warp} feature maps forward. Secondly, feature computation is generally much more expensive than task computation for a range of vision tasks \cite{DFF}, an observation we formalize in Section \ref{sec:IRA}. This, combined with the first idea, suggests the utility of only computing features on select frames, even if we must output segmentations for every frame in a video.

\subsection{Block motion vectors}

MPEG-compressed video consists of two logical components: reference frames, called I-frames, and delta frames, called P-frames. Reference frames are still RGB frames from the video, usually represented as spatially-compressed JPEG images. Delta frames, which introduce temporal compression to video, consist of two subcomponents: block motion vectors and residuals.

Motion vector maps, the artifact of interest in our current work, define a correspondence between pixels in the current frame and pixels in the previous frame. They are generated using \textit{block motion compensation}, a standard component of video compression algorithms \cite{Richardson}:
\begin{enumerate}
	\item Divide the current frame into a non-overlapping grid of 16x16 pixel blocks.
	\item For each block in the current frame, determine the ``best matching" 16x16 block in the previous frame.
	A common matching metric is to minimize the mean squared error between the source block and the target block.
	\item For each block in the current frame, represent the pixel offset to the best matching block in the previous frame as an $(x, y)$ pair.
\end{enumerate}

The resulting grid of $(x, y)$ offsets forms the \textit{block motion vector map} for the current frame. For a $16M \times 16N$ frame, this map has dimensions $M \times N$. The residuals then consist of the pixel-level difference between the current frame, and the previous frame transformed by the motion vectors.

\subsection{Feature Propagation}

Many cameras compress video by default as a means for efficient storage and transmission. The availability of a free form of motion estimation at inference time, the block motion vectors in MPEG-compressed video, suggests the following scheme for fast video segmentation (see \textbf{Algorithm \ref{alg:propagation}}).

Choose a keyframe interval $n$. On keyframes (every $n$\textsuperscript{th} frame), execute the feature network $N_{feat}$ to obtain a feature map. Cache these computed features, $f_c$, and then execute the task network $N_{task}$ on the features to obtain the keyframe segmentation. On intermediate frames, extract the motion vector map $\text{mv}[i]$ corresponding to the current frame index. Warp the cached features $f_c$ one frame forward via bilinear interpolation with $-\text{mv}[i]$. (To warp forward, we apply the negation of the vector map.) Here we employ the differentiable, parameter-free warping operator first proposed by \cite{STN}. Finally, execute $N_{task}$ on the warped features to obtain the current segmentation.

\vspace{-2mm}
\begin{algorithm}
	\caption{Feature propagation}
	\label{alg:propagation}
	\begin{algorithmic}[1]
		\State \textbf{input}: video frames $\{I_{i}\}$, motion vectors $\text{mv}$, keyframe interval $n$
		\For{\text{frame} $I_i$ \textbf{in} $\{I_{i}\}$}
		\If{i \textbf{mod} n = 0}	\Comment{keyframe}
		\State $f_i \leftarrow N_{feat}(I_i)$	 \Comment{keyframe features}
		\State $S_i \leftarrow N_{task}(f_i)$
		\Else{}	 \Comment{intermediate frame}
		\State $f_i \leftarrow \textsc{warp}(f_c, -\text{mv}[i]))$ \Comment{warp cached features}
		\State $S_i \leftarrow N_{task}(f_i)$
		\EndIf
		\State $f_c \leftarrow f_i$						\Comment{cache features}
		\EndFor
		\State \textbf{output}: frame segmentations $\{S_{i}\}$
	\end{algorithmic}
\end{algorithm}
\vspace{-3mm}

Note that warping with a block motion vector map has a natural analog in video decompression. A video decoder takes as input the previous frame and the current block motion vector map, and outputs the current frame (modulo the residuals). Our warp operator takes the previous feature estimate and the current block motion vector map, and outputs the current feature estimate.

\subsection{Inference Runtime Analysis} \label{sec:IRA}

\begin{figure}[t]
	\centering
	\includegraphics[width=0.55\linewidth]{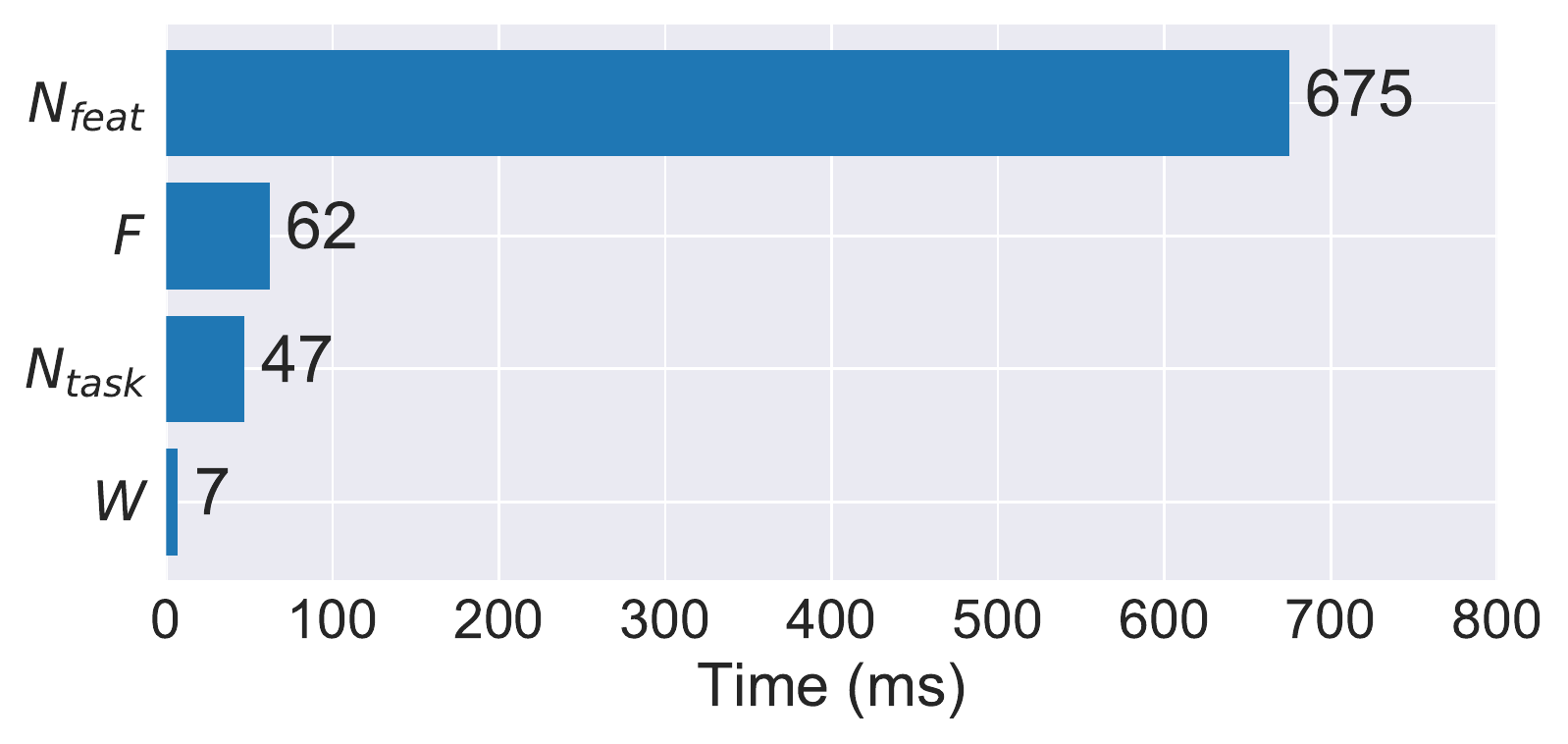}
	\vspace{-2mm}
	\caption{\small A sample runtime breakdown. $F$ is the optical flow net used in \cite{DFF} for motion estimation. $W$ is the warp operator. Models: DeepLab, ResNet-101. GPU: Tesla K80.}
	\label{fig:run}
	\vspace{-2mm}
\end{figure}

Feature propagation is effective because it relegates feature extraction, the most expensive network component, to select keyframes. Of the three remaining operations performed on intermediate frames -- motion estimation, feature warping, and task execution time -- motion estimation with optical flow is the most expensive (see Fig. \ref{fig:run}). By using block motion vectors, we eliminate this remaining bottleneck, accelerating inference times on intermediate frames for a DeepLab segmentation network \cite{DeepLab} from 116 ms per frame ($F + W + N_{task}$) to 54 ms per frame ($W + N_{task}$). For keyframe interval $n$, this translates to a speedup of 53\% on $\frac{n-1}{n}$ of the video frames.

Note that for a given keyframe interval $n$, as we reduce inference time on intermediate frames to zero, we approach a maximum attainable speedup factor of $n$ over a frame-by-frame baseline that runs the full model on every frame. Exceeding this bound, without compromising on accuracy, requires an entirely new approach to feature estimation, the subject of the next section.

We also benchmarked the time required to extract block motion from raw video (i.e. H.264 compression time), and found that ffmpeg takes 2.78 seconds to compress 1,000 Cityscapes video frames, or 2.78 ms per frame. In contrast, optical flow computation on a frame pair takes 62 ms (Fig. \ref{fig:run}). We include this comparison for completeness: since compression is a default behavior on modern cameras, block motion extraction is not a true component of inference time.

\subsection{Feature Interpolation} \label{sec:feat-int}

Given an input video stream, we wish to compute the segmentation of every frame as efficiently as possible, while preserving accuracy. In a batch setting, we have access to the entire video, and desire the segmentations for all the frames, as input to another model (e.g. an autonomous control system). In a streaming setting, we have access frames as they come in, but may be willing to tolerate a small delay of keyframe interval $n$ frames ($\frac{n}{30}$ seconds or $33.3n$ milliseconds at 30 fps) before we output a segmentation, if that means we can match the throughput of the video stream and maintain high accuracy.

We make two observations. First, all intermediate frames in a video by definition lie between two designated keyframes, which represent bounds on the current scene. New objects that are missed in forward feature propagation schemes are more likely to be captured if both past and incoming keyframes are used. Second, feature fusion techniques are effective at preserving strong signals in any one input feature map, as seen in \cite{TwoStream16}. This suggests the viability of estimating intermediate frame features as the fusion of the features of enclosing keyframes.

\begin{figure}[t]
	\centering
	\vspace{-2mm}
	\includegraphics[width=9cm]{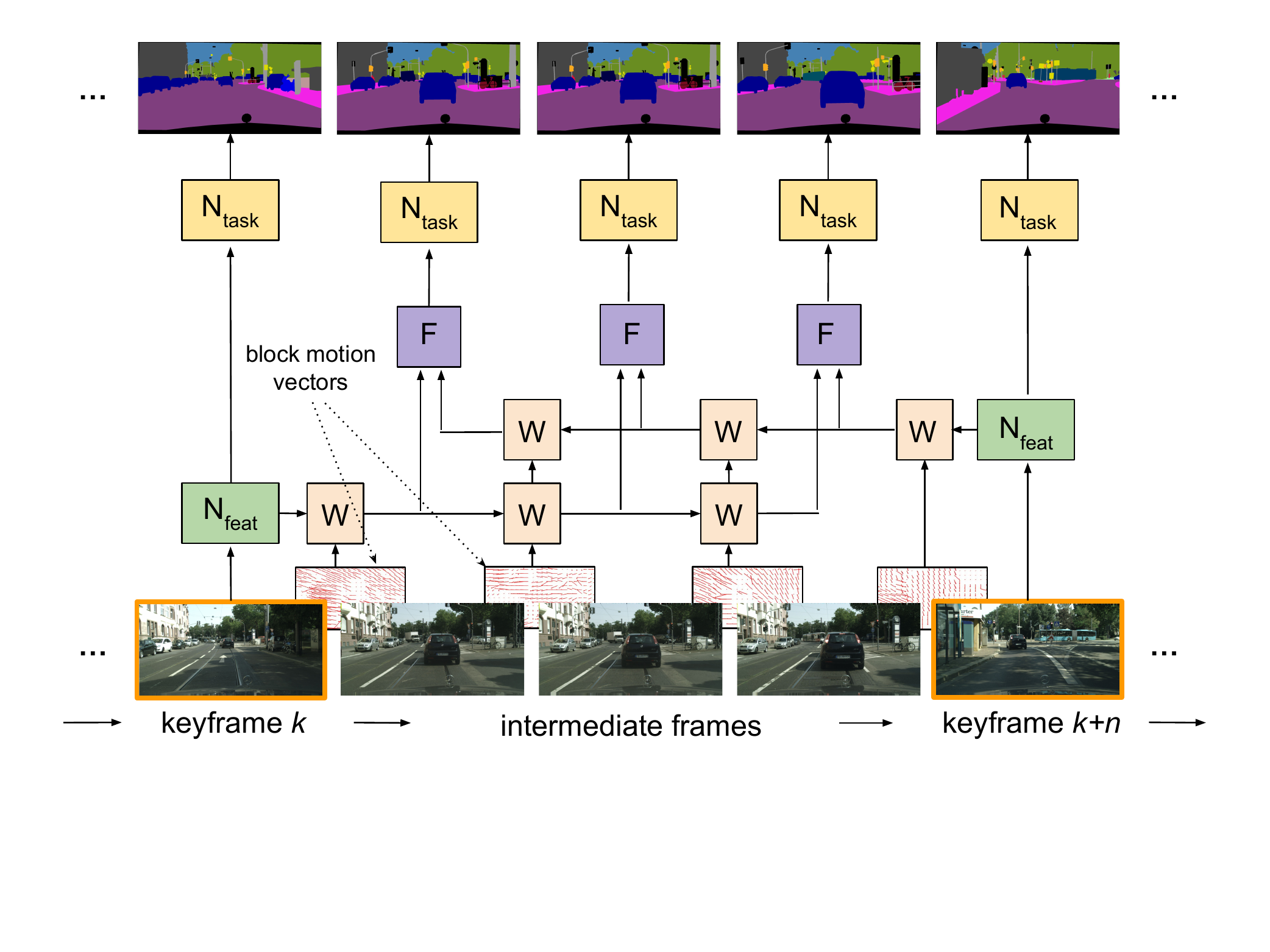}
	\vspace{-16mm}
	\caption{In feature interpolation, features for intermediate frames are estimated as a fusion ($F$) of features warped ($W$) inward from enclosing keyframes.}
	\label{fig:interp}
	\vspace{-4mm}
\end{figure}

Expanding on this idea, we propose \textbf{Algorithm \ref{alg:interpolation}}. On any given keyframe, precompute the features for the \textit{next} keyframe. On intermediate frames, warp the previous keyframe's features, $N_{feat}(I_{k})$, forward to the current frame $I_i$ using incremental forward motion estimates, $-\text{mv}[k: i]$. Warp the next keyframe's features, $N_{feat}(I_{k+n})$, \textit{backward} to the current frame using incremental backward motion estimates, $\text{mv}[k+n : i]$. Fuse the two feature maps using either a simple weighted average or a learned fusion operator, $F$. Then execute the task network $N_{task}$ on the fused feature maps (see Fig. \ref{fig:interp}).

To eliminate redundant computation, on keyframes, we \textit{precompute} forward and backward warped feature maps $f^f, f^b$ corresponding to each subsequent intermediate frame, $\{I_{k+1}, ..., I_{k+n-1}\}$. For keyframe interval $n$, this amounts to $n-1$ forward warped feature maps and $n-1$ backward warped feature maps.

We expect this scheme to outperform basic feature propagation for two reasons. First, assuming fusion is beneficial, the fused feature maps will yield higher accuracy segmentations than using the forward or backward propagated feature maps alone. (In feature propagation, we only have access to the former.) Second, by propagating feature maps from both the previous and subsequent keyframes, we reduce the distance the closer keyframe's features must be warped by a factor of two. For a given keyframe interval, in feature propagation, keyframe features are warped $\frac{n-1}{2}$ steps in expectation. In interpolation, the closer keyframe's features are warped half as many steps in expectation. As a result, when features are warped in both directions, even if we were to just select the closer feature map (instead of fusing the two), we would have a better estimate of the current frame's features. Since warping and fusion are significantly cheaper than other network components, this scheme improves accuracy at little additional inference cost. We validate both observations in Section \ref{sec:exp}.

\begin{algorithm}
	\caption{Feature interpolation}
	\label{alg:interpolation}
	\begin{algorithmic}[1]
		\State \textbf{input}: video frames $\{I_{i}\}$, motion vectors $\text{mv}$, keyframe interval $n$
		\State $W_{f}, W_{b} \leftarrow []$		\Comment{forward, backward warped features}
		\For{\text{frame} $I_i$ \textbf{in} $\{I_{i}\}$}
		\If{i \textbf{mod} n == 0}	\Comment{keyframe}
		\State $f_i \leftarrow N_{feat}(I_i)$	 \Comment{curr keyframe features}
		\State $f_{i+n} \leftarrow N_{feat}(F_{i+n})$	 \Comment{next keyframe features}
		\State $W_{f} \leftarrow \textsc{propagate}(f_{i}, n - 1, -\text{mv}[i+1: i+n])$
		\State $W_{b} \leftarrow \textsc{propagate}(f_{i+n}, n - 1, \text{mv}[i+n: i+1])$
		\Else{}	 \Comment{intermediate frame}
		\State $p \leftarrow \text{i \textbf{mod} n}$	\Comment{offset from prev keyframe}
		\State $f_i \leftarrow F(\frac{n-p}{n} \cdot W_f[\text{p}], \frac{p}{n} \cdot W_b[n - \text{p}])$ \Comment{fuse propagated features}
		\EndIf
		\State $S_i \leftarrow N_{task}(f_i)$
		\EndFor
		\State \textbf{output}: frame segmentations $\{S_{i}\}$
		\newline
		\Function{propagate}{features $f$, steps $n$, warp array $g$} \Comment{warp $f$ for $n$ steps}
		\State $O \leftarrow [f]$
		\For{i = 1 \textbf{to} $n$}
		\State $\text{append}(O, \textsc{warp}(O[i-1], g[i]))$		\Comment{warp features one step}
		\EndFor
		\State \Return $O$
		\EndFunction
	\end{algorithmic}
\end{algorithm}
\vspace{-1mm}

\subsection{Feature Fusion}

We consider several possible fusion operators: max fusion, average fusion, and convolutional fusion. We implement max and average fusion by aligning the input feature maps $f^f, f^b \in R^{1 \times C \times h \times w}$ along the channel dimension, and computing a maximum or average across each pixel in corresponding channels, a parameter-free operation. We implement convolutional fusion by \textit{stacking} the input feature maps along the channel dimension $[f^f, f^b]_C = f^{s} \in R^{1 \times 2C \times h \times w}$, and applying a bank of learned, $1 \times 1$ convolutional filters and biases that reduce the input channel dimensionality by a factor of two. In practice, we repurpose the first layer of the task network, which is itself a $1 \times 1$ conv layer, into our fusion layer, to avoid adding to the total inference time.

Before applying the fusion operator at inference time, we weight the two input feature maps $f^f, f^b$ by scalars $\alpha$ and $1 - \alpha$, respectively, that correspond to \textit{feature relevance}, a scheme that works very well in practice. For keyframe interval $n$, and a frame at offsets $p$ and $n - p$ from the previous and next keyframes, respectively, we set $\alpha = \frac{n - p}{n}$ and $1 - \alpha = \frac{p}{n}$, thereby penalizing the input features warped \textit{farther }from their keyframe. Thus, when $p$ is small relative to $n$, we weight the previous keyframe's features more heavily, and vice versa. As an example, when $n=5$ and $p=1$, we set $\alpha = \frac{4}{5}$ and $1 - \alpha = \frac{1}{5}$. To summarize, the features for intermediate frame $I_i$ are set to: $f_i = F(\frac{n-p}{n} f^f, \frac{p}{n} f^b)$, where $p = i \text{ mod } n$. This weighting procedure is reflected in Algorithm \ref{alg:interpolation}.

\section{Experiments} \label{sec:exp}

\subsection{Setup}

\subsubsection{Datasets.} We train and evaluate our system on Cityscapes \cite{Cityscapes} and CamVid \cite{CamVid}, two popular, large-scale datasets for complex urban scene understanding. Cityscapes consists of 30-frame video snippets shot at 17 fps and $2048 \times 1024$ pixels. CamVid consists of 10 minutes of footage captured at 30 fps and $960 \times 720$ pixels. Ground-truth labels are provided for the 20\textsuperscript{th} frame in each Cityscapes \texttt{train} and \texttt{val} split snippet, and for every 30\textsuperscript{th} frame in the CamVid dataset. On Cityscapes, we train on the \texttt{train} split and evaluate on the \texttt{val} split, following the example of previous work \cite{DRN, DeepLab, DFF, DCN}. On CamVid, we adopt the standard train-test split of Sturgess, et al \cite{Sturgess}. We use the standard mean intersection-over-union (mIoU) metric to evaluate semantic segmentation accuracy, and measure throughput in frames per second (fps) to evaluate inference performance.

\subsubsection{Architecture.} For our segmentation network, we adopt a variant of the DeepLab architecture called Deformable DeepLab \cite{DCN}, which employs \textit{deformable convolutions} in the last ResNet block (conv5) to achieve significantly higher accuracy at comparable inference cost to a standard DeepLab model. DeepLab \cite{DeepLab} is widely considered a state-of-the-art architecture for semantic segmentation, and a DeepLab implementation currently ranks first on the PASCAL VOC object segmentation challenge \cite{Pascal}. Our DeepLab model uses ResNet-101 as its feature network, which produces intermediate representations $f_i \in R^{1 \times 2048 \times \frac{h}{16} \times \frac{w}{16}}$. The DeepLab task network outputs predictions $p_i \in R^{1 \times C \times h \times w}$, where $C$ is 12 or 20 for the number of object classes in CamVid and Cityscapes, respectively.

\subsubsection{Training.} To train our single-frame DeepLab model, we initialize with weights from an ImageNet-trained ResNet-101 model, and learn task-specific weights on the Cityscapes and CamVid \texttt{train} sets. To train our video segmentation system, we sample at random a labeled image from the train set, and select a preceding and succeeding frame to serve as the previous and next keyframe, respectively. Since motion estimation with block motion vectors and feature warping are both parameter-free, feature propagation introduces no additional weights. Training feature interpolation with convolutional fusion, however, involves learning weights for the $1 \times 1$ conv fusion layer, which is applied to stacked feature maps, each with channel dimension $2048$.

In all cases, we train with stochastic gradient descent on an AWS EC2 instance with 4 Tesla K80 GPUs for 50 epochs, starting with a learning rate of $10^{-3}$ if learning any weights from scratch and $10^{-4}$ if fine-tuning.

\subsubsection{Evaluation.} To evaluate our video segmentation system on a sparsely annotated image dataset, we utilize the following setup. At inference time, we select an operational keyframe interval $i$, and iterate over the test examples, choosing keyframes such that the distance to the labeled frame rotates uniformly between $0$ and $i-1$. This sampling procedure simulates evaluation on a densely labeled video dataset, where $\frac{1}{i}$ frames fall at each keyframe offset between $0$ and $i-1$.

\subsection{Results}

\subsubsection{Baseline.} For our performance and accuracy baseline, we evaluate our full DeepLab model on every labeled frame in the Cityscapes and CamVid \texttt{test} splits. This frame-by-frame evaluation procedure forms the conventional approach to semantic segmentation on video. Our baseline model achieves accuracy 75.2 mIoU on Cityscapes, matching published results for the DeepLab architecture we used \cite{DCN}, and a throughput of 1.3 frames per second (fps). On CamVid, the baseline model achieves 68.6 mIoU at a throughput of 3.7 frames per second.

\subsubsection{Propagation and Interpolation.} In this section, we evaluate our two main contributions: 1) feature propagation with block motion vectors (\textbf{prop-mv}), and 2) feature interpolation, our new feature estimation scheme, implemented with block motion vectors (\textbf{interp-mv}). We compare to the closest available prior work on the problem, a feature propagation scheme based on optical flow \cite{DFF} (\textbf{prop-flow}).

\begin{figure}%
	\vspace{-8mm}
	\subfloat[\textbf{Cityscapes}. Data from Table \ref{tbl:cityscapes}.]{\includegraphics[width=6.1cm]{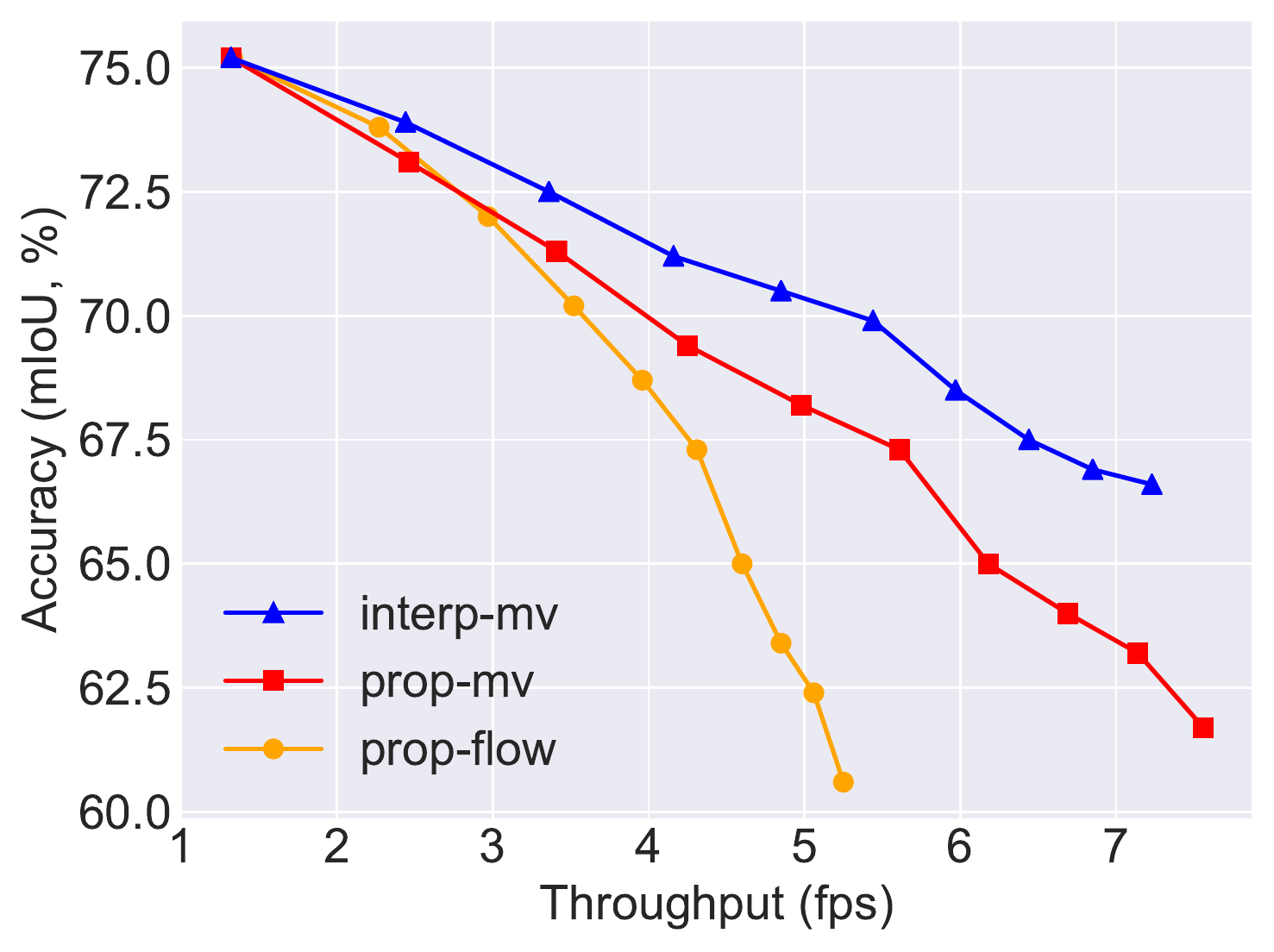} \label{fig:city-avg}}
	\subfloat[\textbf{CamVid}. Data from Table \ref{tbl:camvid} (Appx).]{\includegraphics[width=5.95cm]{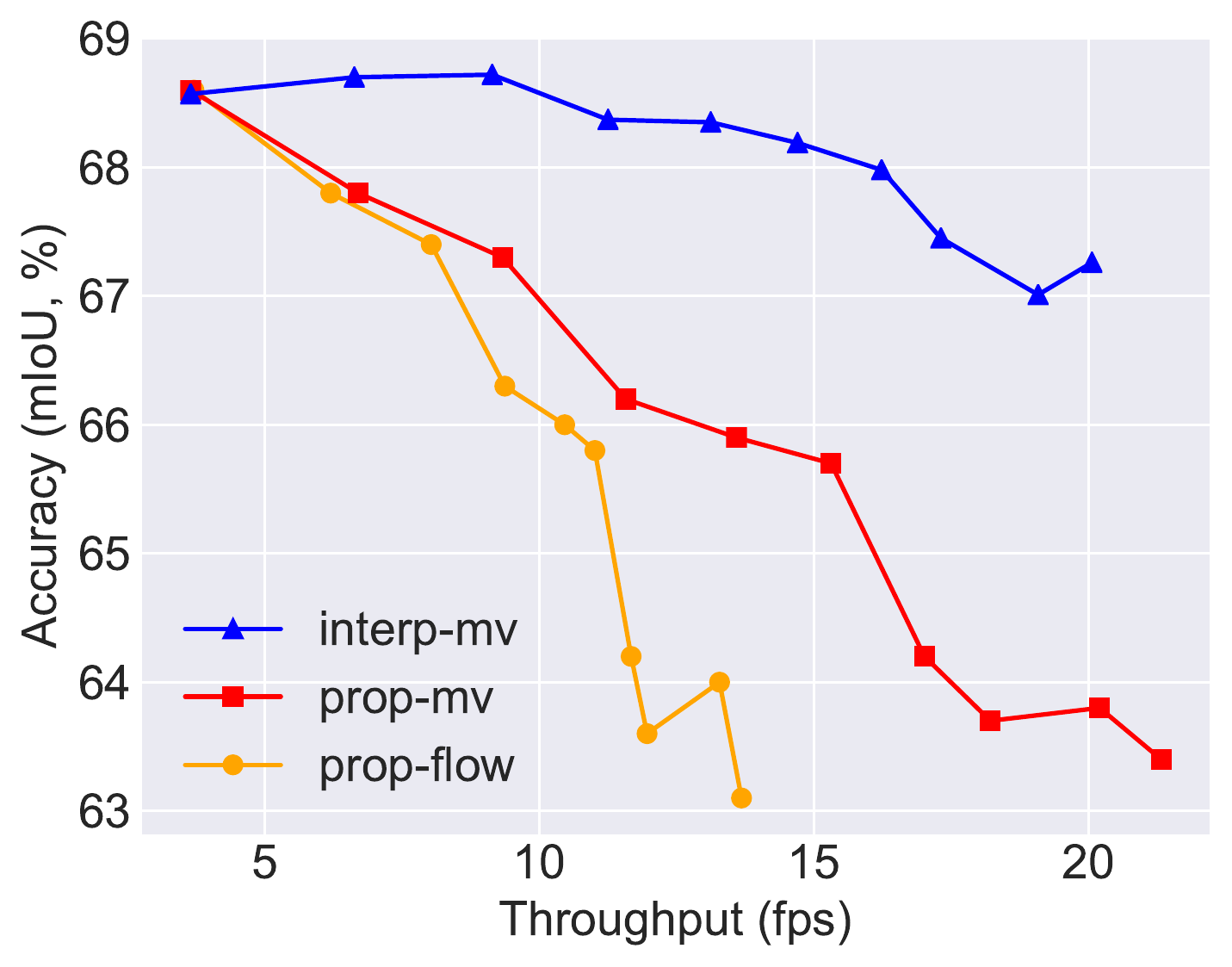} \label{fig:camvid-avg}}%
	\caption{Accuracy (\textbf{avg.}) vs. throughput for all schemes on Cityscapes and CamVid.}%
	\label{fig:acc-avg}%
	\vspace{-4mm}
\end{figure}

We evaluate by comparing accuracy-runtime curves for the three approaches on Cityscapes and CamVid (see \textbf{Figure \ref{fig:acc-avg}}). These curves are generated by plotting accuracy against throughput at each keyframe interval in Table \ref{tbl:cityscapes} and Appendix: Table \ref{tbl:camvid}, which contain comprehensive results. A scheme that allows operation at higher accuracy at every throughput than another approach is said to \textit{strictly outperform} it.

\renewcommand{\arraystretch}{1.2}
\begin{table}[ht]
	\vspace{-2mm}
	\caption{Accuracy and throughput on \textbf{Cityscapes} for three schemes: (1) optical-flow based feature propagation \cite{DFF} (\textbf{prop-flow}), (2) motion vector-based feature propagation (\textbf{prop-mv}), and (3) motion vector-based feature interpolation (\textbf{interp-mv}).}
	\centering
	\label{tbl:cityscapes}
	\begin{tabular}{@{\extracolsep{4pt}}llcccccccccc}
		\toprule   
		{} & {} & \multicolumn{10}{c}{keyframe interval} \\
		\cmidrule{3-12} 
		Metric & Scheme & 1 & 2 & 3 & 4 & 5 & 6 & 7 & 8 & 9 & 10\\ 
		\midrule
		mIoU (\textbf{avg.}) & prop-flow & 75.2 & 73.8 & 72.0 & 70.2 & 68.7 & 67.3 & 65.0 & 63.4 & 62.4 & 60.6\\
		(\%) & prop-mv & 75.2 & 73.1 & 71.3 & 69.4 & 68.2 & 67.3 & 65.0 & 64.0 & 63.2 & 61.7 \\ 
		& interp-mv &  \textbf{75.2} & \textbf{73.9} & \textbf{72.5} & \textbf{71.2} & \textbf{70.5} & \textbf{69.9} & \textbf{68.5} & \textbf{67.5} & \textbf{66.9} & \textbf{66.6} \\
		\midrule
		mIoU (\textbf{min.}) & prop-flow & 75.2 & 72.4 & 68.9 & 65.6 & 62.4 & 59.1 & 56.3 & 54.4 & 52.5 & 50.5 \\
		(\%)  & prop-mv & 75.2 & 71.3 & 67.7 & 64.8 & 62.4 & 60.1 & 58.5 & 56.9 & 55.0 & 53.7\\
		& interp-mv &  \textbf{75.2} & \textbf{72.5} & \textbf{71.5} & \textbf{68.0} & \textbf{67.2} & \textbf{66.2} & \textbf{65.4} & \textbf{64.6} & \textbf{63.5} & \textbf{62.9} \\
		\midrule
		throughput & prop-flow & 1.3 & 2.3 & 3.0 & 3.5 & 4.0 & 4.3 & 4.6 & 4.9 & 5.1 & 5.3 \\ 
		(fps) & prop-mv & 1.3 & 2.5 & 3.4 & 4.3 & 5.0 & 5.6 & 6.2 & 6.7 & 7.1 & 7.6 \\ 
		& interp-mv & 1.3 & 2.4 & 3.4 & 4.2 & 4.9 & 5.4 & 6.0 & 6.4 & 6.9 & 7.2 \\ 
		\bottomrule
	\end{tabular}
	\vspace{-3mm}
\end{table}
\renewcommand{\arraystretch}{1.0}

First, we note that block motion-based feature propagation (\textbf{prop-mv}) \textit{outperforms} optical flow-based propagation (prop-flow) at all but the lowest throughputs. While motion vectors are slightly less accurate than optical flow in general, by cutting inference times by 53\% on intermediate frames (Sec. \ref{sec:IRA}), prop-mv allows operation at much lower keyframe intervals than optical flow to achieve the same inference speeds. This effect is pronounced enough to result in a much more favorable accuracy-throughput trade-off curve for prop-mv.

Second, we find that our feature interpolation scheme (\textbf{interp-mv}) \textit{strictly outperforms} both feature propagation schemes. At every keyframe interval, interp-mv is more accurate than prop-flow and prop-mv; moreover, it operates at similar throughput to prop-mv. This translates to a consistent advantage over prop-mv, and an even larger advantage over prop-flow (see Fig. \ref{fig:acc-avg}).

These trends are even more pronounced on the CamVid dataset (see Fig. \ref{fig:camvid-avg} and Appendix). Since CamVid consists of smaller images, shot at a higher frame rate, prop-mv and interp-mv achieve faster runtime speeds (up to 20+ fps) and lower accuracy degradation (less than 6\%) than on Cityscapes. In particular, interp-mv actually registers a small \textit{accuracy gain} over the baseline at keyframe intervals 2 and 3, utilizing multi-frame context to improve on the accuracy of the single-frame DeepLab model, in addition to achieving faster inference times.

\textbf{Metrics.} We also note a distinction between two metrics: the standard \textit{average} accuracy, the results for which are plotted in Fig. \ref{fig:acc-avg}, and \textit{minimum} accuracy, which is a measure of the lowest frame-level accuracy an approach entails, i.e. accuracy on frames farthest away from keyframes (Appendix: Fig. \ref{fig:acc-min}). Minimum accuracy is the appropriate metric to consider when we wish to ensure that all frame segmentations are at least as accurate as some threshold value.

In particular, consider a batch processing setting in which the goal is to segment a video as efficiently as possible, at an accuracy target of no less than 65 mIoU on any frame, as calibrated at training time. As Table \ref{tbl:cityscapes} demonstrates, at that accuracy threshold, feature interpolation enables operation at 6.0 fps on Cityscapes. This is significantly faster than achievable inference speeds with feature propagation alone, using either optical flow (3.5 fps) or block motion vectors (4.3 fps). In general, feature interpolation achieves almost twice the throughput as \cite{DFF} on Cityscapes (\textit{more} than twice on CamVid), at any target accuracy.

\textbf{Baseline.} We also compare to our frame-by-frame DeepLab baseline, which offers low throughput but high average accuracy. As Table \ref{tbl:cityscapes} indicates, even at average accuracy about 70 mIoU on Cityscapes, a figure competitive with the best single-frame models \cite{DRN, DeepLab, DCN}, feature interpolation offers speedups of $2-4\times$ over the baseline. On CamVid, the gains are even larger: at keyframe interval 10, interpolation achieves a $5.6\times$ speedup over the baseline at just 1.3\% lower mIoU. In fact, at keyframe interval 3, interpolation obtains a $2.5\times$ speedup over the baseline, at slightly \textit{higher} accuracy.

\textbf{Delay.} Finally, recall that to use feature interpolation, we must accept a delay of keyframe interval $n$ frames, which corresponds to $\frac{n}{30}$ seconds at 30 fps. For example, at $n = 5$, interpolation introduces a delay of $\frac{5}{30}$ seconds, or $167$ ms. By comparison, prop-flow [7] takes 250 ms to segment a frame at key interval 5, and interp-mv takes 200 ms. Thus, by lagging by less than 1 segmentation, we are able to segment $3.8\times$ more frames per hour than the frame-by-frame model (5 fps vs. 1.3 fps). This is a suitable tradeoff in almost all batch settings (e.g. segmenting 1000s of hours of video to generate training data for a driverless vehicle; post-hoc surveillance video analysis), and in interactive applications such as video anomaly detection and film editing. Note that operating at a higher keyframe interval introduces a longer delay, but also enables much higher throughput.

\begin{figure}%
	\vspace{-1mm}
	\includegraphics[width=2.95cm]{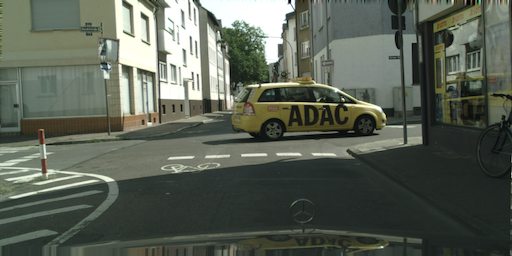} \hfill
	\includegraphics[width=2.95cm]{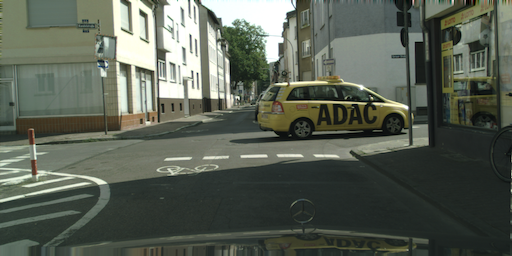} \hfill
	\includegraphics[width=2.95cm]{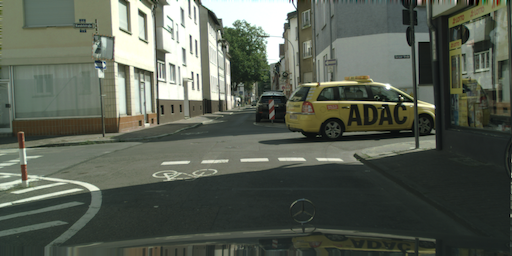} \hfill
	\includegraphics[width=2.95cm]{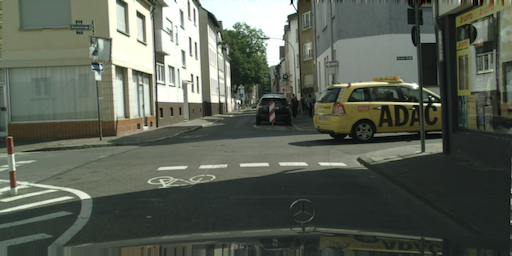} \\[1pt]
	\includegraphics[width=2.95cm]{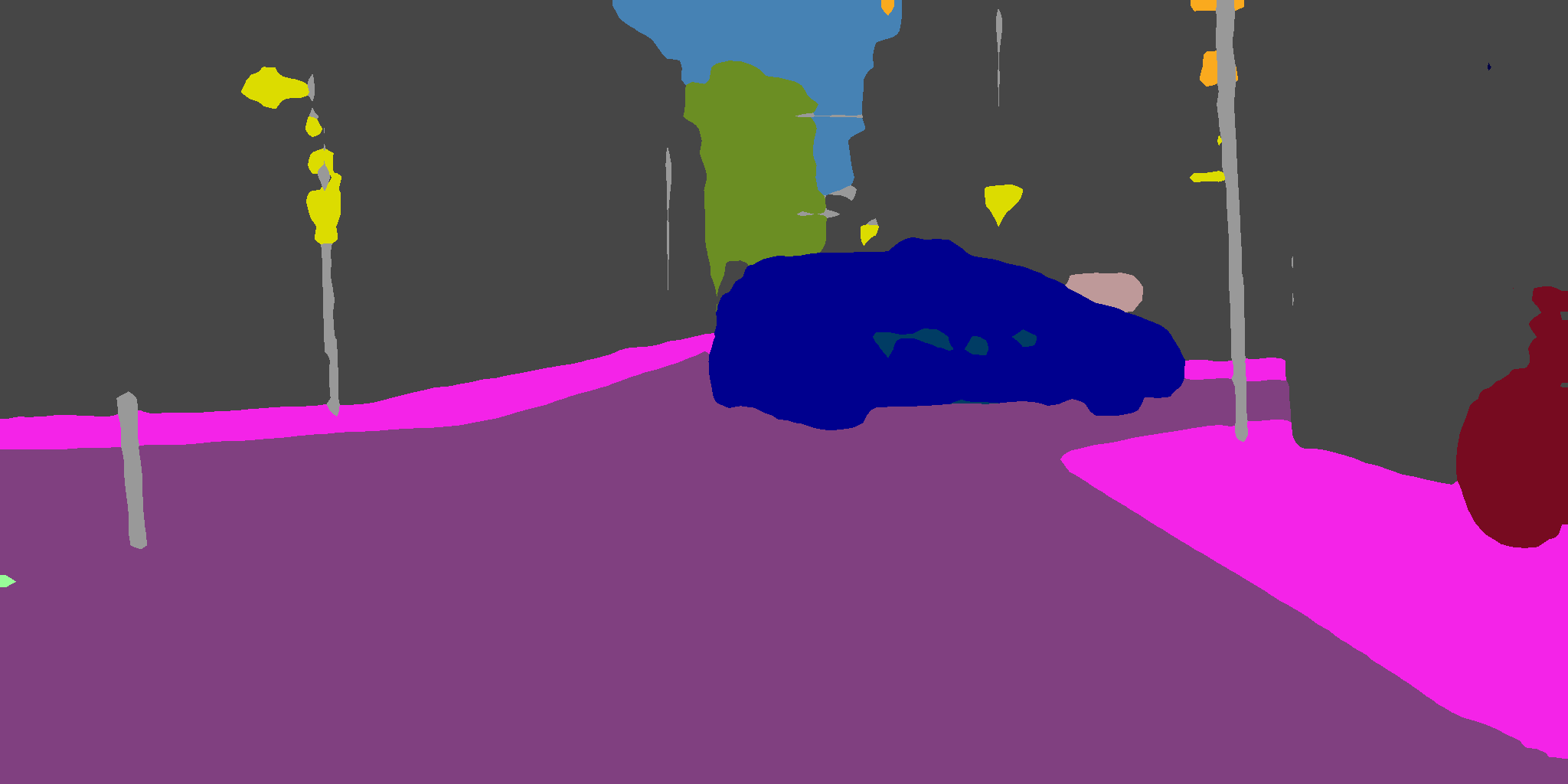} \hfill
	\includegraphics[width=2.95cm]{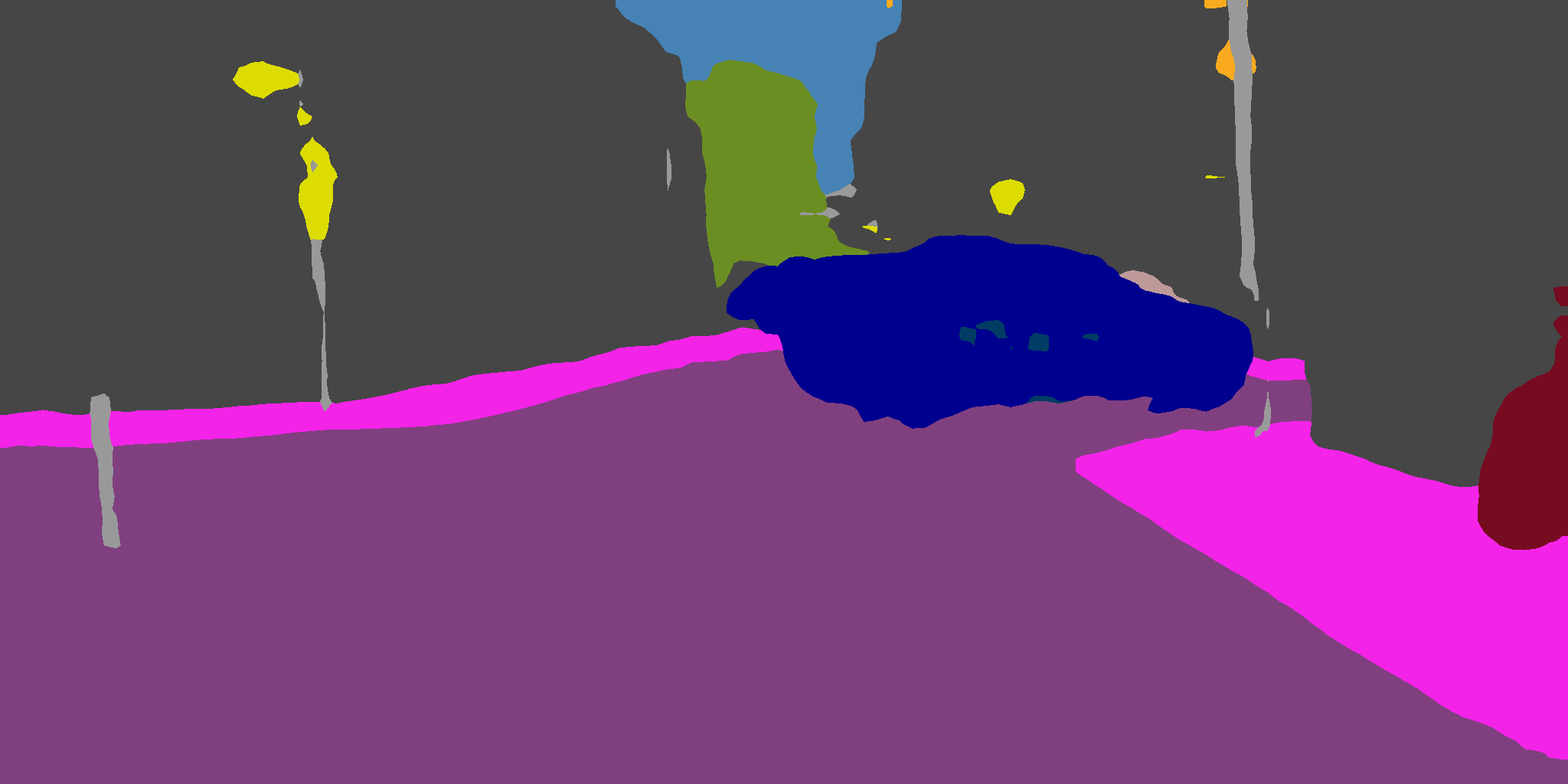} \hfill
	\includegraphics[width=2.95cm]{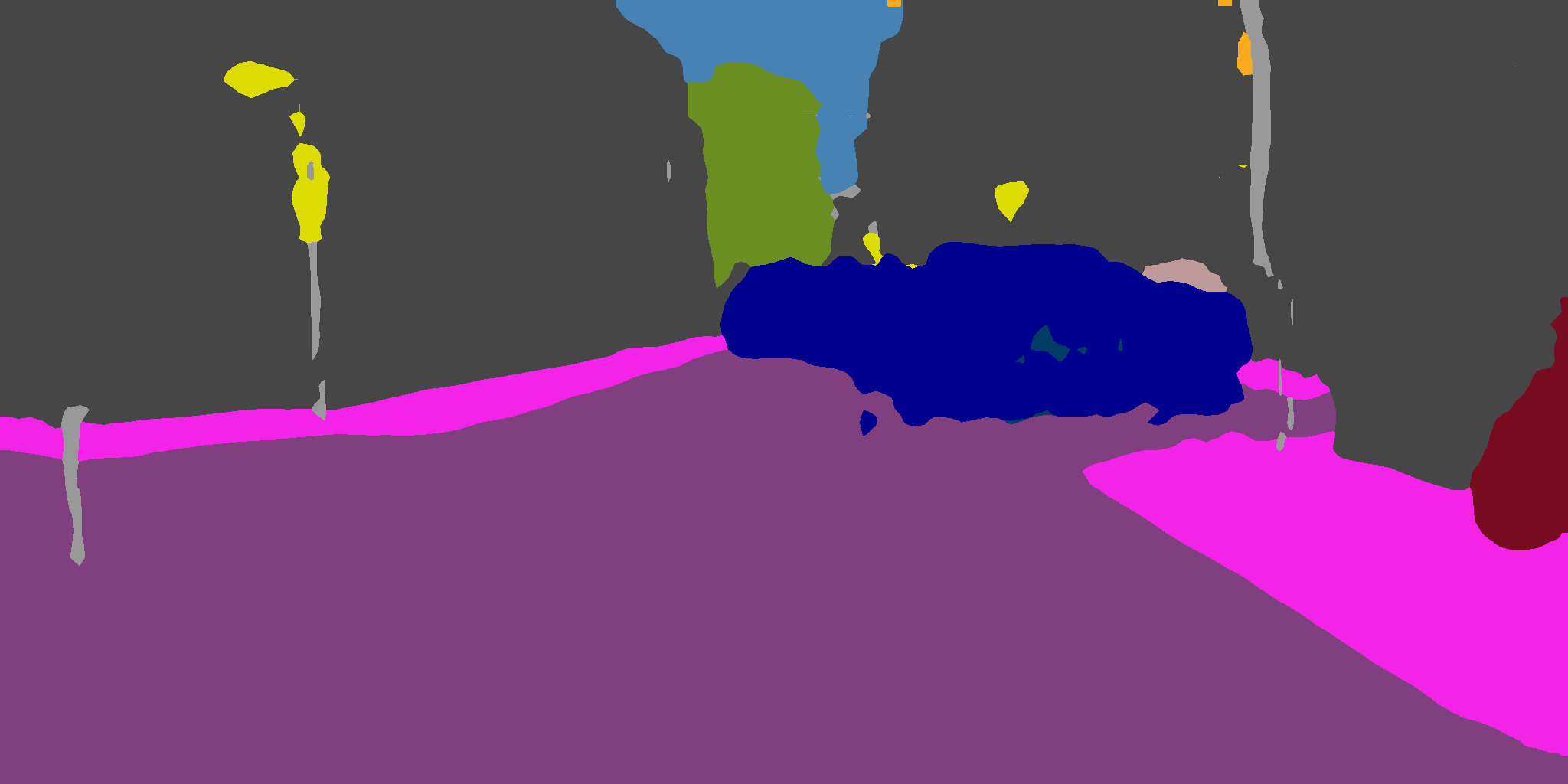} \hfill
	\includegraphics[width=2.95cm]{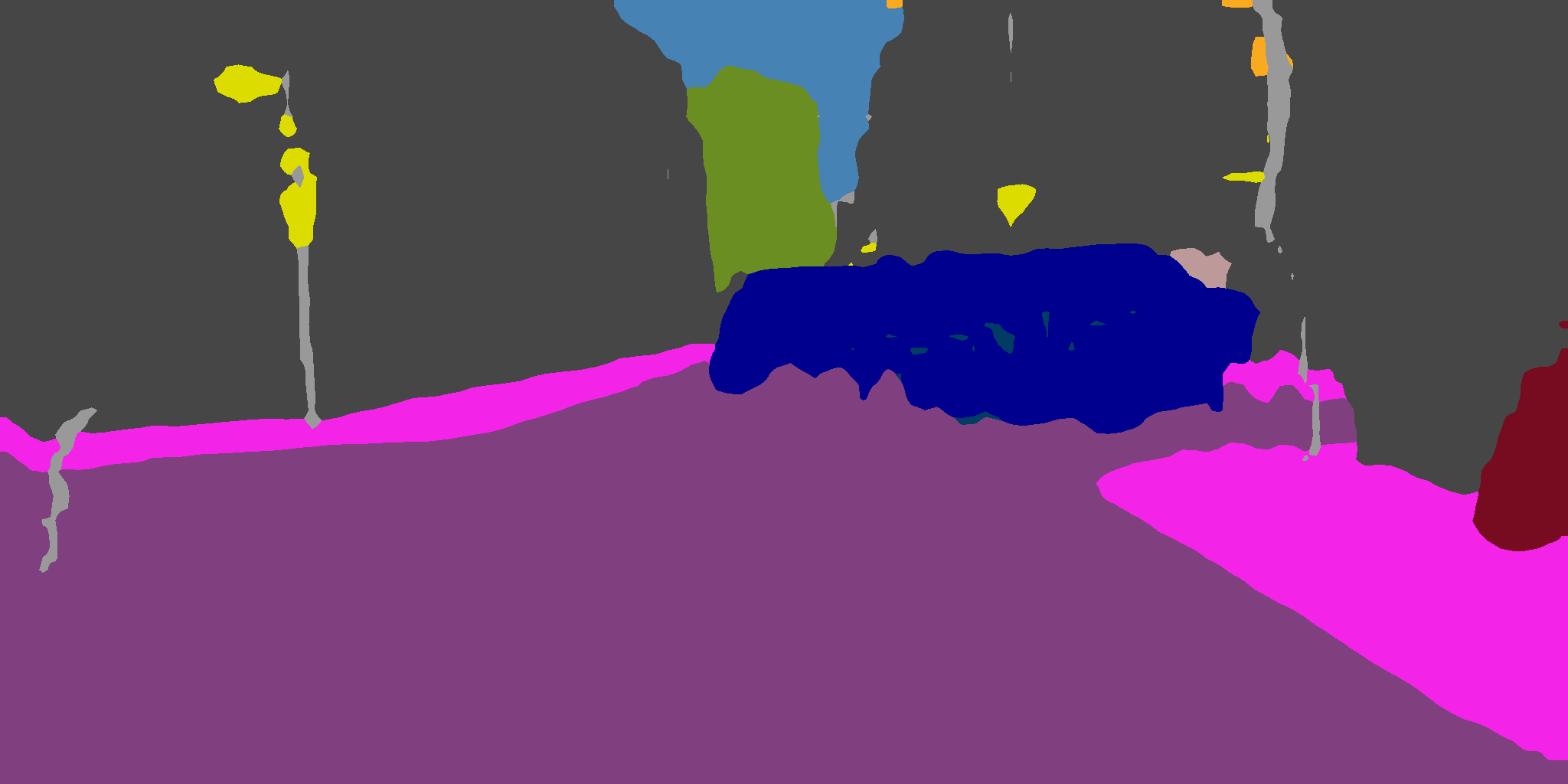} \\[-8pt]
	\subfloat[k     ]{\includegraphics[width=2.95cm]{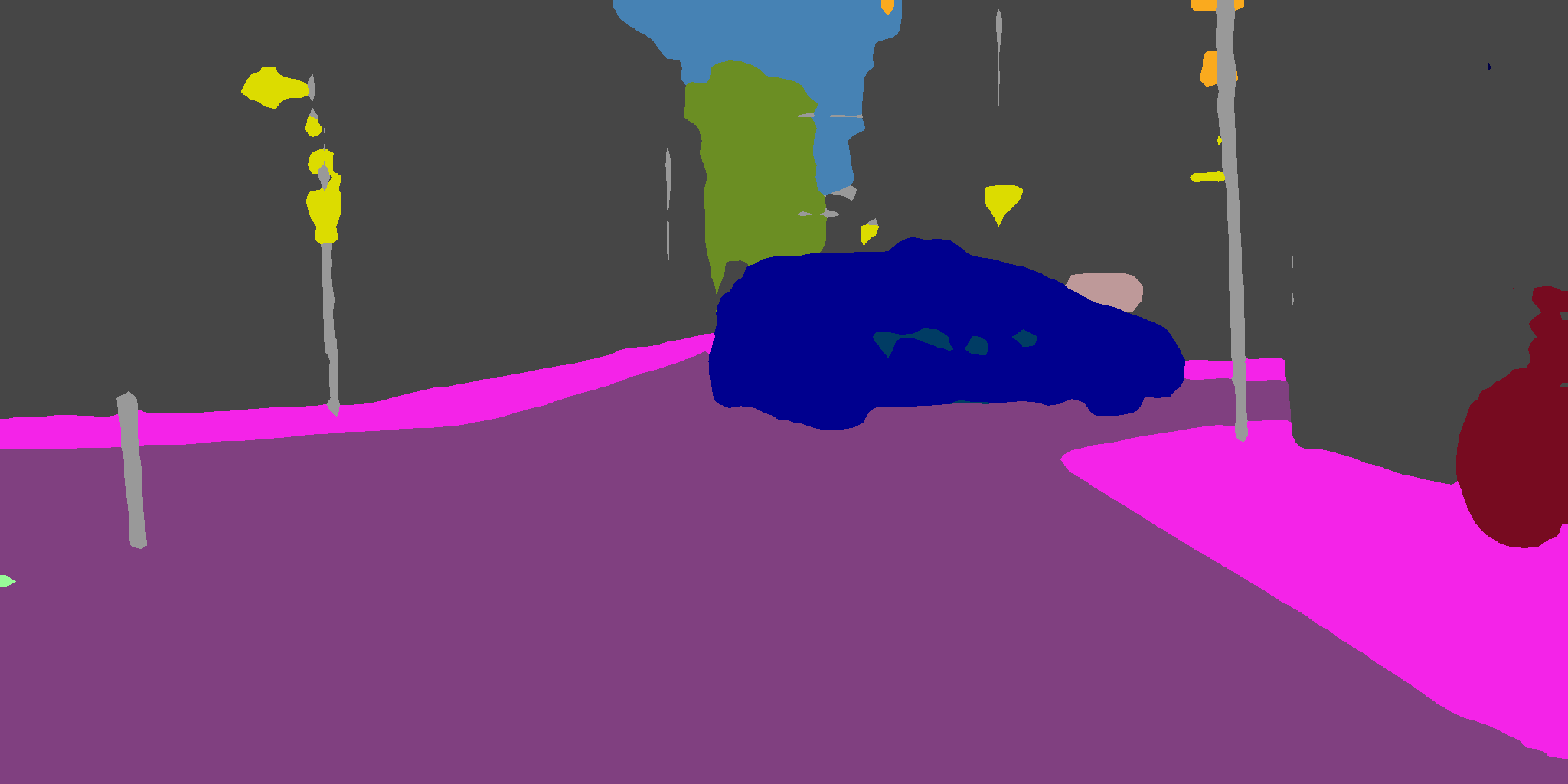}} \hfill
	\subfloat[k+2]{\includegraphics[width=2.95cm]{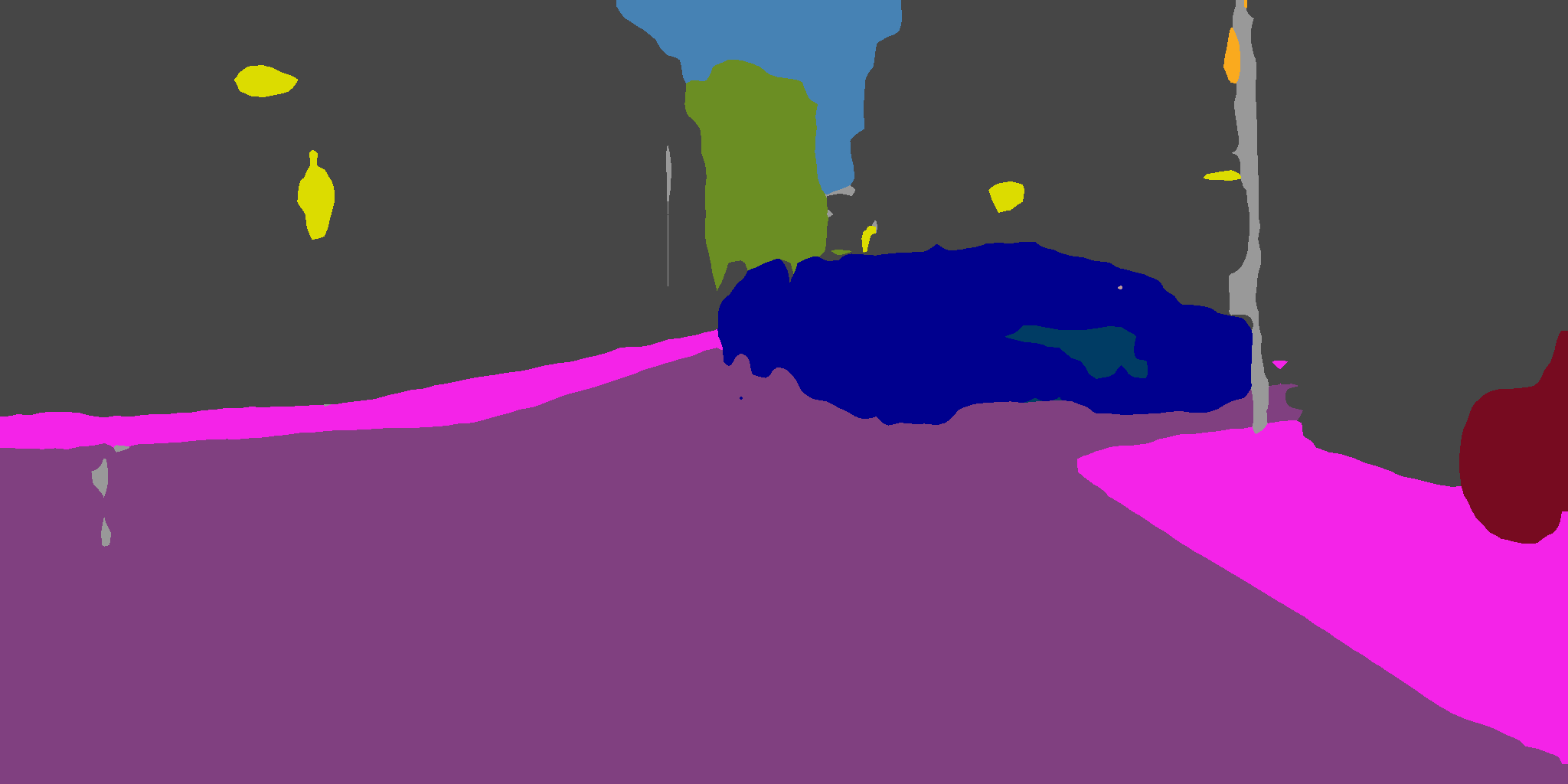}} \hfill
	\subfloat[k+4]{\includegraphics[width=2.95cm]{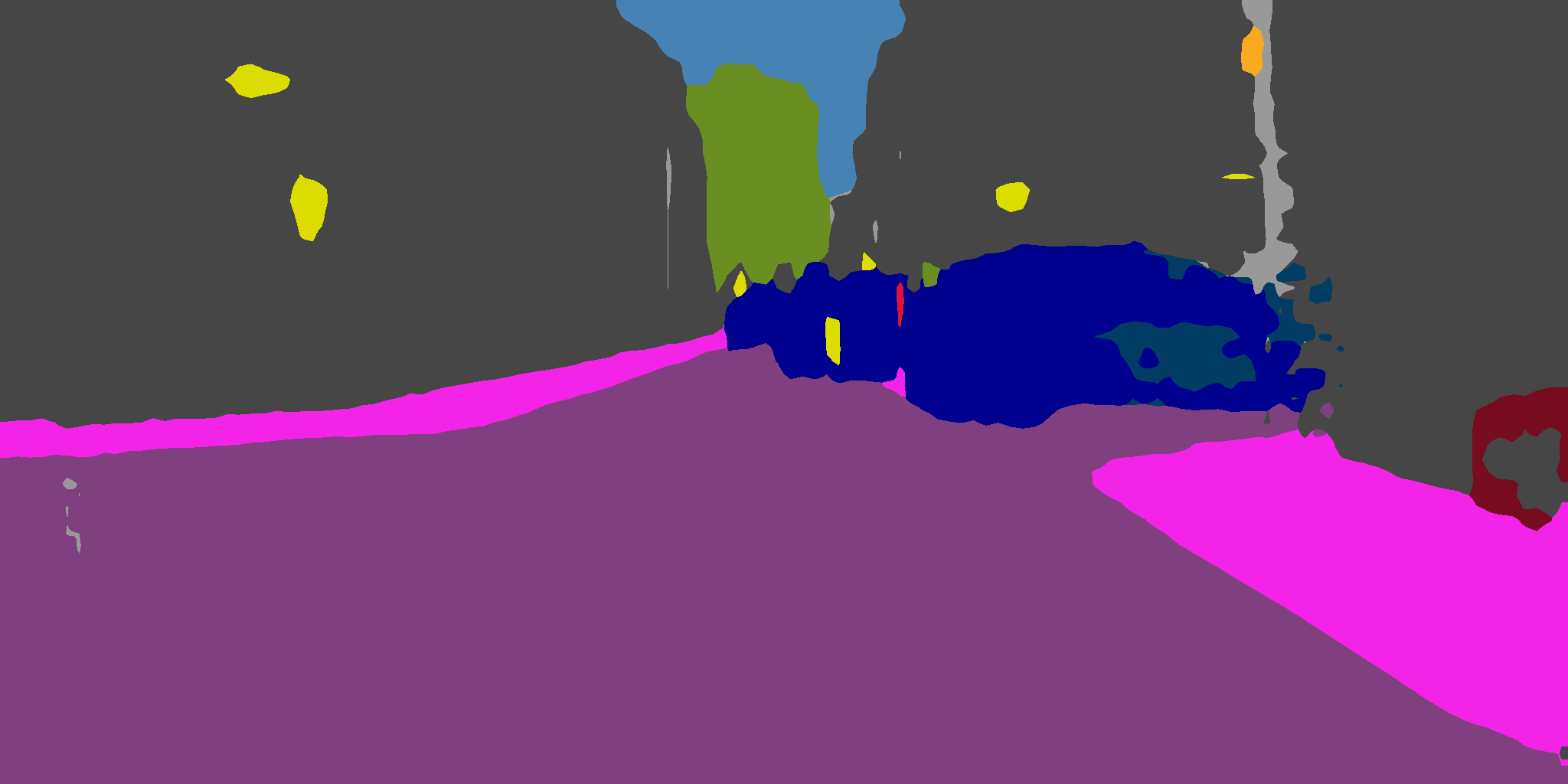}} \hfill
	\subfloat[k+6]{\includegraphics[width=2.95cm]{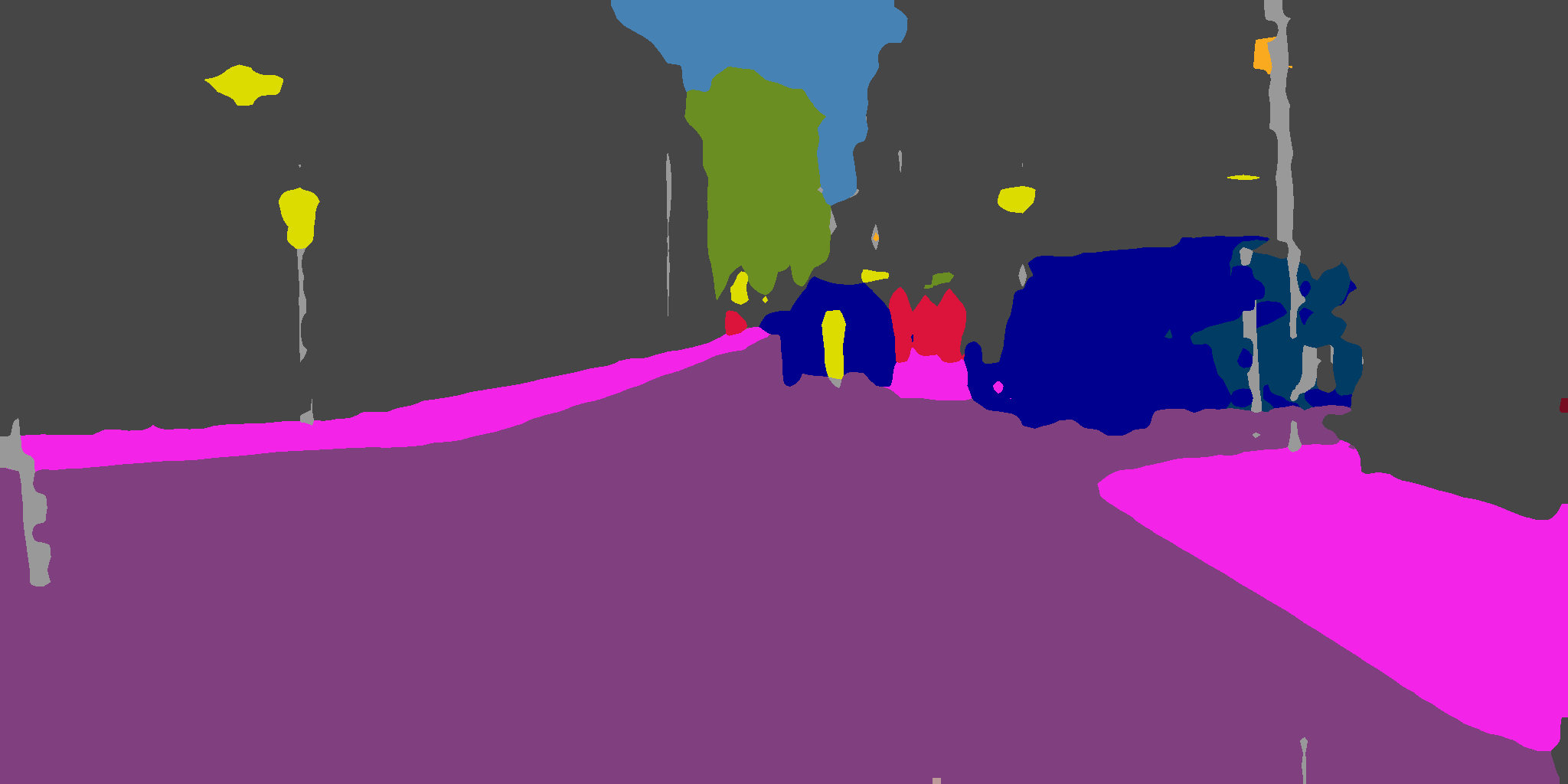}}
	\caption{Example segmentations at keyframe interval $7$. Column $k+i$ corresponds to outputs $i$ frames past the selected keyframe $k$. \textbf{First row:} input frames. \textbf{Second row:} prop-flow \cite{DFF}. \textbf{Third row:} interp-mv (us). Note that, by $k+6$, prop-flow has significant warped the moving car, obscuring the people, vehicle, and street sign in the background (image center), while these entities remain clearly visible with interpolation. \textit{Cityscapes}.}%
	\label{fig:qual-out}%
	\vspace{-4mm}
\end{figure}

Fig. \ref{fig:qual-out} depicts a \textbf{qualitative comparison} of interpolation and prop-flow \cite{DFF}.

\subsubsection{Feature Fusion.} In this second set of experiments, we evaluate the accuracy gain achieved by feature fusion, in order to isolate the contribution of feature fusion to the success of our feature interpolation scheme. As Table \ref{tbl:fusion} demonstrates, utilizing any fusion strategy, whether max, average, or conv fusion, results in higher accuracy than using either input feature map alone. This holds true even when one feature map is significantly stronger than the other (rows 2-4), and for both short and long distances to the keyframes. This observed additive effect suggests that feature fusion is highly effective at capturing signal that appears in only one input feature map, and in merging spatial information across time.

\renewcommand{\arraystretch}{1.2}
\begin{table}[ht]
	\vspace{-1mm}
	\caption{An evaluation of feature fusion. We report final accuracies for various keyframe placements. Forward and Backward refer to the input feature maps (and their direction of warping). \textit{Cityscapes.}}
	\centering
	\label{tbl:fusion}
	\begin{tabular}{@{\extracolsep{4pt}}lccccc}
		\toprule   
		Distance & Forward & Backward & Max Fusion & Avg. Fusion & Conv. Fusion \\
		to keyframe(s) & mIoU & mIoU & mIoU & mIoU & mIoU \\
		\midrule
		1 & 71.8 & 69.9 & 72.6 & 72.8 & 72.6 \\
		2 & 67.8 & 62.4 & 68.2 & 68.5 & 68.2 \\
		3 & 64.9 & 59.8 & 66.3 & 66.7 & 66.4 \\
		4 & 62.4 & 57.3 & 64.5 & 65.0 & 64.7 \\
		\midrule
	\end{tabular}
	\vspace{-8mm}
\end{table}
\renewcommand{\arraystretch}{1.0}

\section{Conclusion}

In this paper, we develop two main contributions: 1) a feature propagation scheme that uses block motion vectors from compressed video (e.g. H.264 codecs) to warp features from frame-to-frame cheaply and accurately, and 2) a new feature estimation scheme for video, termed feature interpolation, that utilizes feature propagation, frame information from the near future, and feature fusion, to produce accurate segmentations at high throughput. We evaluate on the Cityscapes and CamVid datasets, and find that our schemes enable significant speedups, at any accuracy, over both a frame-by-frame baseline and prior work. Our methods are general, and represent an important advance in the effort to operate image models efficiently on video.

\bibliographystyle{splncs04}
\bibliography{egbib}

\begin{thebibliography}{10}
\providecommand{\url}[1]{\texttt{#1}}
\providecommand{\urlprefix}{URL }
\providecommand{\doi}[1]{https://doi.org/#1}

\bibitem{Pascal}
Aytar, Y.: Pascal voc challenge performance evaluation and download server.
  \url{http://host.robots.ox.ac.uk:8080/leaderboard}, accessed: 2018-03-06

\bibitem{SegNet}
Badrinarayanan, V., Kendall, A., Cipolla, R.: Segnet: A deep convolutional
  encoder-decoder architecture for image segmentation. In: PAMI (2017)

\bibitem{CamVid}
Brostow, G.J., Fauqueur, J., Cipolla, R.: Semantic object classes in video: A
  high-definition ground truth database. Pattern Recognition Letters
  \textbf{30}(2),  88–97 (2009)

\bibitem{DeepLabV1}
Chen, L.C., Papandreou, G., Kokkinos, I., Murphy, K., Yuille, A.L.: Semantic
  image segmentation with deep convolutional nets and fully connected crfs. In:
  ICLR (2016)

\bibitem{DeepLab}
Chen, L.C., Papandreou, G., Kokkinos, I., Murphy, K., Yuille, A.L.: Deeplab:
  Semantic image segmentation with deep convolutional nets, atrous convolution,
  and fully connected crfs. In: PAMI (2017)

\bibitem{Cityscapes}
Cordts, M., Omran, M., Ramos, S., Rehfeld, T., Enzweiler, M., Benenson, R.,
  Franke, U., Roth, S., Schiele, B.: The cityscapes dataset for semantic urban
  scene understanding. In: CVPR (2016)

\bibitem{DCN}
Dai, J., Qi, H., Xiong, Y., Li, Y., Zhang, G., Hu, H., Wei, Y.: Deformable
  convolutional networks. In: ICCV (2017)

\bibitem{FlowNet}
Dosovitskiy, A., Fischer, P., Ilg, E., H{\"a}usser, P., Hazrbas, C., Golkov,
  V., v.d. Smagt, P., Cremers, D., Brox, T.: Flownet: Learning optical flow
  with convolutional networks. In: ICCV (2015)

\bibitem{STResNets}
Feichtenhofer, C., Pinz, A., Wildes, R.: Spatiotemporal residual networks for
  video action recognition. In: NIPS (2016)

\bibitem{DeepReps}
Feichtenhofer, C., Pinz, A., Wildes, R.P., Zisserman, A.: What have we learned
  from deep representations for action recognition? In: CVPR (2018)

\bibitem{TwoStream16}
Feichtenhofer, C., Pinz, A., Zisserman, A.: Convolutional two-stream network
  fusion for video action recognition. In: CVPR (2016)

\bibitem{MovingObjects}
Fragkiadaki, K., Arbelaez, P., Felsen, P., Malik, J.: Learning to segment
  moving objects in videos. In: CVPR (2015)

\bibitem{NetWarp}
Gadde, R., Jampani, V., Gehler, P.V.: Semantic video cnns through
  representation warping. In: ICCV (2017)

\bibitem{FlowNet2}
Ilg, E., Mayer, N., Saikia, T., Keuper, M., Dosovitskiy, A., Brox, T.: Flownet
  2.0: Evolution of optical flow estimation with deep networks. In: CVPR (2017)

\bibitem{STN}
Jaderberg, M., Simonyan, K., Zisserman, A., Kavukcuoglu, K.: Spatial
  transformer networks. In: NIPS (2015)

\bibitem{FusionSeg}
Jain, S.D., Xiong, B., Grauman, K.: Fusionseg: Learning to combine motion and
  appearance for fully automatic segmentation of generic objects in videos. In:
  CVPR (2017)

\bibitem{TCNN}
Kang, K., Li, H., Yan, J., Zeng, X., Yang, B., Xiao, T., Zhang, C., Wang, Z.,
  Wang, R., Wang, X., Ouyang, W.: T-cnn: Tubelets with convolutional neural
  networks for object detection from videos. In: CVPR (2016)

\bibitem{Karpathy}
Karpathy, A., Toderici, G., Shetty, S., Leung, T., Sukthankar, R., Fei-Fei, L.:
  Large-scale video classification with convolutional neural networks. In: CVPR
  (2014)

\bibitem{FSO}
Kundu, A., Vineet, V., Koltun, V.: Feature space optimization for semantic
  video segmentation. In: CVPR (2016)

\bibitem{RefineNet}
Lin, G., Milan, A., Shen, C., Reid, I.: Refinenet: Multi-path refinement
  networks for high-resolution semantic segmentation. In: CVPR (2017)

\bibitem{FCN}
Long, J., Shelhamer, E., Darrell, T.: Fully convolutional networks for semantic
  segmentation. In: CVPR (2015)

\bibitem{BudgetAware}
Mahasseni, B., Todorovic, S., Fern, A.: Budget-aware deep semantic video
  segmentation. In: CVPR (2017)

\bibitem{FewStrokes}
Nagaraja, N.S., Schmidt, F.R., Brox, T.: Video segmentation with just a few
  strokes. In: CVPR (2015)

\bibitem{FastObjSeg}
Papazoglou, A., Ferrari, V.: Fast object segmentation in unconstrained video.
  In: ICCV (2013)

\bibitem{Richardson}
Richardson, I.E.: H.264 and MPEG-4 video compression: video coding for
  next-generation multimedia. Wiley (2008)

\bibitem{CC}
Shelhamer, E., Rakelly, K., Hoffman, J., Darrell, T.: Clockwork convnets for
  video semantic segmentation. In: Video Semantic Segmentation Workshop at ECCV
  (2016)

\bibitem{TwoStream14}
Simonyan, K., Zisserman, A.: Two-stream convolutional networks for action
  recognition in videos. In: NIPS (2014)

\bibitem{Sturgess}
Sturgess, P., Alahari, K., Ladick\'y, L., Torr, P.H.S.: Combining appearance
  and structure from motion features for road scene understanding. In: BMVC
  (2009)

\bibitem{ObjectFlow}
Tsai, Y.H., Yang, M.H., Black, M.J.: Video segmentation via object flow. In:
  CVPR (2016)

\bibitem{SceneFlow}
Vedula, S., Baker, S., Rander, P., Collins, R., Kanade, T.: Three-dimensional
  scene flow. In: ICCV (1999)

\bibitem{PointTrack}
Veenman, C.J., Reinders, M.J.T., Backer, E.: Motion tracking as a constrained
  optimization problem. Pattern Recognition  \textbf{36},  2049--2067 (2003)

\bibitem{ActiveFrame}
Vijayanarasimhan, S., Grauman, K.: Active frame selection for label propagation
  in videos. In: ECCV (2012)

\bibitem{CoViAR}
Wu, C.Y., Zaheer, M., Hu, H., Manmatha, R., Smola, A.J., Krähenbühl, P.:
  Compressed video action recognition. In: CVPR (2018)

\bibitem{Multiscale}
Yu, F., Koltun, V.: Multi-scale context aggregation by dilated convolutions.
  In: ICLR (2016)

\bibitem{DRN}
Yu, F., Koltun, V., Funkhouser, T.: Dilated residual networks. In: CVPR (2017)

\bibitem{MVecCNNs}
Zhang, B., Wang, L., Wang, Z., Qiao, Y., Wang, H.: Real-time action recognition
  with enhanced motion vector cnns. In: CVPR (2016)

\bibitem{PSPNet}
Zhao, H., Shi, J., Qi, X., Wang, X., Jia, J.: Pyramid scene parsing network.
  In: CVPR (2017)

\bibitem{DFF2}
Zhu, X., Dai, J., Yuan, L., Wei, Y.: Toward high performance video object
  detection. In: CVPR (2018)

\bibitem{DFF}
Zhu, X., Xiong, Y., Dai, J., Yuan, L., Wei, Y.: Deep feature flow for video
  recognition. In: CVPR (2017)

\end{thebibliography}

\clearpage

\section{Appendix}

\subsection{Results}

\begin{figure}%
	\vspace{-8mm}
	\subfloat[\textbf{Cityscapes}. Data from Table \ref{tbl:cityscapes}.]{\includegraphics[width=6.1cm]{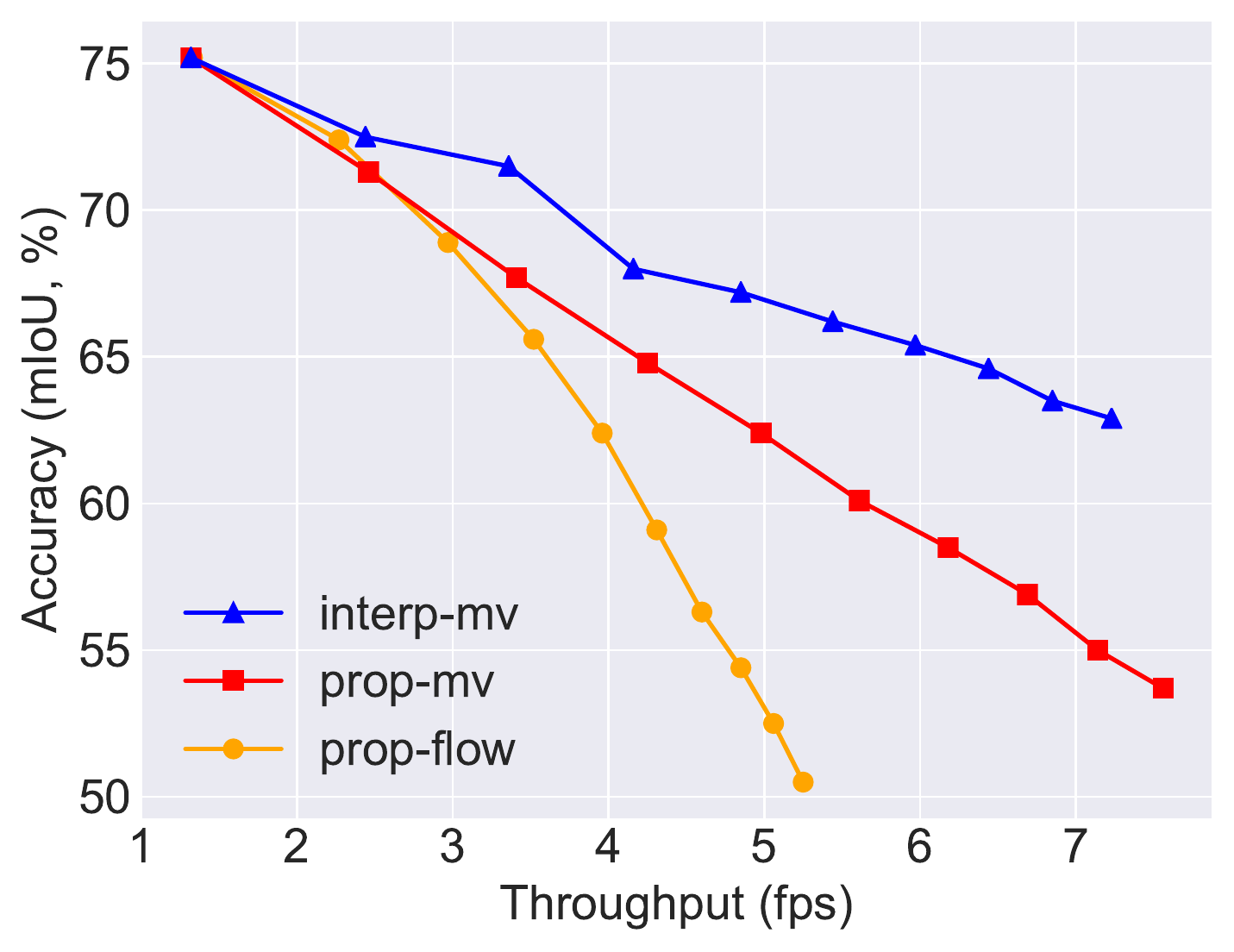} \label{fig:cityscapes-min}}
	\subfloat[\textbf{CamVid}. Data from Table \ref{tbl:camvid}.]{\includegraphics[width=6.1cm]{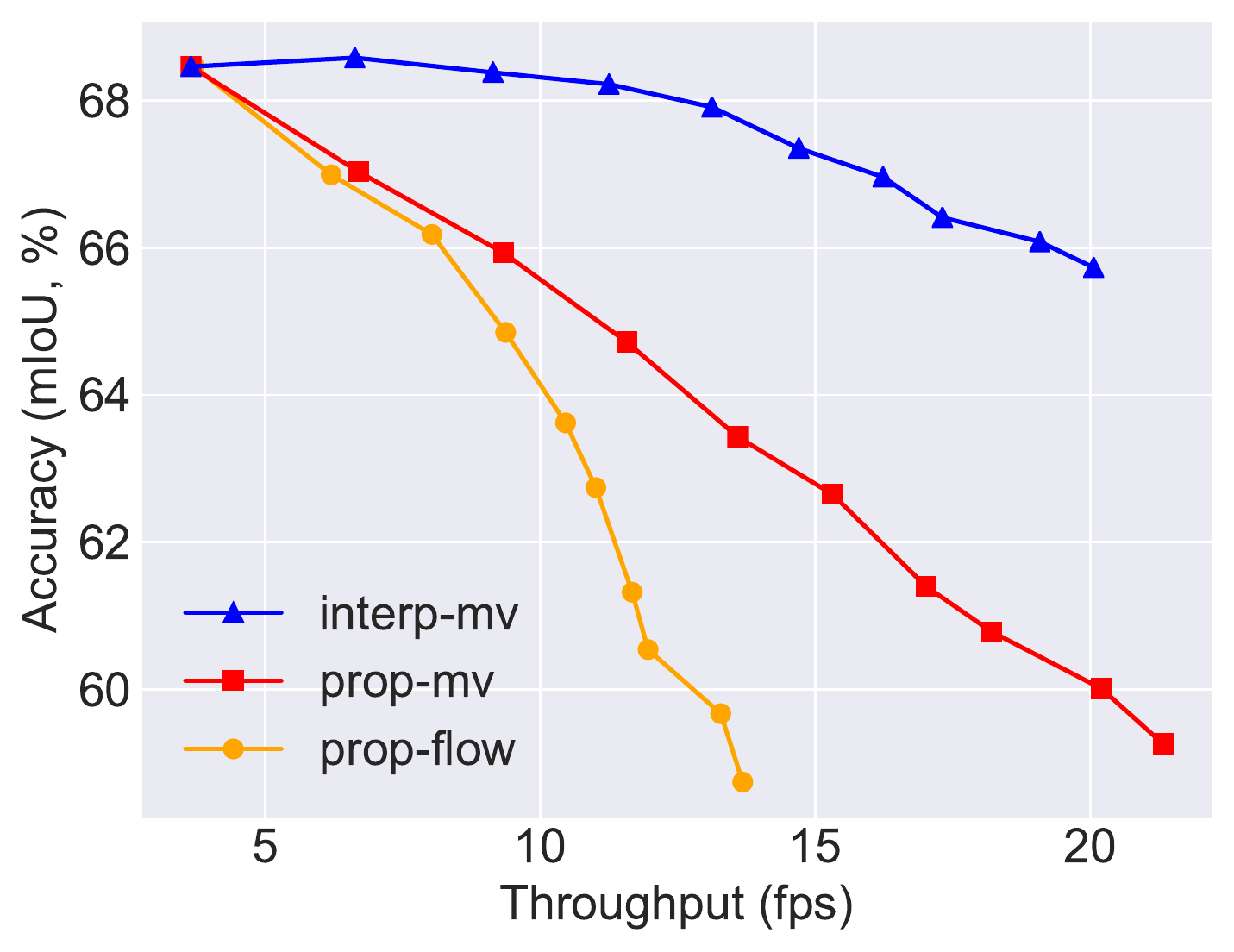} \label{fig:camvid-min}}%
	\caption{Accuracy (\textbf{min.}) vs. throughput for all schemes on Cityscapes and CamVid.}%
	\label{fig:acc-min}%
	\vspace{-2mm}
\end{figure}

Results on the CamVid dataset (\textbf{Table \ref{tbl:camvid}}) and minimum accuracy vs. throughput plots for Cityscapes and CamVid (\textbf{Figure \ref{fig:acc-min}}).

\renewcommand{\arraystretch}{1.2}
\begin{table}[ht]
	\vspace{-0mm}
	\caption{Accuracy and throughput on the \textbf{CamVid} dataset for the three schemes: prop-flow \cite{DFF}, prop-mv, and interp-mv.}
	\centering
	\label{tbl:camvid}
	\begin{tabular}{@{\extracolsep{4pt}}llcccccccccc}
		\toprule
		{} & {} & \multicolumn{10}{c}{keyframe interval} \\
		\cmidrule{3-12}
		Metric & Scheme & 1 & 2 & 3 & 4 & 5 & 6 & 7 & 8 & 9 & 10\\
		\midrule
		mIoU (\textbf{avg.}) & prop-flow & 68.6 & 67.8 & 67.4 & 66.3 & 66.0 & 65.8 & 64.2 & 63.6 & 64.0 & 63.1 \\
		(\%) & prop-mv & 68.6 & 67.8 & 67.3 & 66.2 & 65.9 & 65.7 & 64.2 & 63.7 & 63.8 & 63.4  \\
		& interp-mv & \textbf{68.6} & \textbf{68.7} & \textbf{68.7} & \textbf{68.4} & \textbf{68.4} & \textbf{68.2} & \textbf{68.0} & \textbf{67.5} & \textbf{67.0} & \textbf{67.3} \\
		\midrule
		mIoU (\textbf{min.}) & prop-flow & 68.5 & 67.0 & 66.2 & 64.9 & 63.6 & 62.7 & 61.3 & 60.5 & 59.7 & 58.7 \\
		(\%)  & prop-mv & 68.5 & 67.0 & 65.9 & 64.7 & 63.4 & 62.7 & 61.4 & 60.8 & 60.0 & 59.3 \\
		& interp-mv & \textbf{68.5} & \textbf{68.6} & \textbf{68.4} & \textbf{68.2} & \textbf{67.9} & \textbf{67.4} & \textbf{67.0} & \textbf{66.4} & \textbf{66.1} & \textbf{65.7} \\
		\midrule
		throughput & prop-flow & 3.6 & 6.2 & 8.0 & 9.4 & 10.5 & 11.0 & 11.7 & 12.0 & 13.3 & 13.7 \\
		(fps) & prop-mv & 3.6 & 6.7 & 9.3 & 11.6 & 13.6 & 15.3 & 17.0 & 18.2 & 20.2 & 21.3 \\
		& interp-mv & 3.6 & 6.6 & 9.1 & 11.3 & 13.1 & 14.7 & 16.2 & 17.3 & 19.1 & 20.1 \\
		\bottomrule
	\end{tabular}
	\vspace{-4mm}
\end{table}
\renewcommand{\arraystretch}{1.0}

On CamVid, interp-mv achieves \textit{well over twice} the throughput as prop-flow, at any minimum accuracy threshold (see Fig. \ref{fig:camvid-min}). For example, at accuracy target 66 mIoU, interp-mv enables operation at 19.1 fps, compared to only 8.0 fps with prop-flow. At accuracy target 67 mIoU, interp-mv enables 16.2 fps, compared to 6.2 fps with prop-flow. This trend holds over the entire domain.

\end{document}